\documentclass[10pt,journal,compsoc]{IEEEtran}
\usepackage{arydshln}
\usepackage{epsfig}
\usepackage{graphicx}
\usepackage{amsmath}
\usepackage{amssymb}
\usepackage{subfigure}
\usepackage{algorithm}
\usepackage{algorithmic}
\usepackage{multirow}
\usepackage{booktabs}
\usepackage{xcolor}
\usepackage{amsthm}
\usepackage{wrapfig}

\usepackage{booktabs}       
\usepackage{amsfonts}       
\usepackage{nicefrac}       
\usepackage{microtype}      
\usepackage{xcolor}         
\usepackage{epsfig}
\usepackage{graphicx}
\usepackage{amsmath}
\usepackage{amssymb}
\usepackage{booktabs}
\usepackage{multirow}
\usepackage{arydshln}
\usepackage{graphicx}
\usepackage{amsmath}
\usepackage{amssymb}
\usepackage{booktabs}
\usepackage{multirow}
\usepackage{algorithm}
\usepackage{algorithmic}
\usepackage{xcolor}
\usepackage{graphicx}
\usepackage{arydshln}
\usepackage{graphicx}
\usepackage{makecell}
\usepackage{enumitem}
 \usepackage{wrapfig}

\usepackage{ragged2e} 
\usepackage[colorlinks, citecolor=blue]{hyperref}

\ifCLASSOPTIONcompsoc
  \usepackage[nocompress]{cite}
\else
  \usepackage{cite}
\fi
\ifCLASSINFOpdf
\else
\fi
\definecolor{OpenAIgreen}{RGB}{16,163,127}
\definecolor{bittersweet}{rgb}{1.0, 0.44, 0.37}
\begin{document}
%
\title{Geometry-to-Image Synthesis-Driven Generative Point Cloud Registration}

\author{ Haobo Jiang, Jin Xie, Jian Yang, Liang Yu, and Jianmin Zheng$^{*}$
	\IEEEcompsocitemizethanks{
	        \IEEEcompsocthanksitem *Corresponding author.
			\IEEEcompsocthanksitem Haobo Jiang and Jianmin Zheng are with ANGEL CorpLab and College of Computing and Data Science, Nanyang Technological University, Singapore. \protect\\
			E-mail: \{haobo.jiang, ASJMZheng\}@ntu.edu.sg
			\protect\\

        \IEEEcompsocthanksitem Liang Yu is with Alibaba Cloud, Alibaba Group, China.  \protect\\
		E-mail: liangyu.yl@alibaba-inc.com
		\protect\\

        \IEEEcompsocthanksitem Jian Yang is with PCA Lab, VCIP, College of Computer Science, Nankai University, China.  \protect\\
		E-mail: csjyang@nankai.edu.cn
		\protect\\

  		\IEEEcompsocthanksitem Jin Xie is with State Key Laboratory for Novel Software Technology \& Schoolof Intelligence Science and Technology, Nanjing University, China.  \protect\\
		E-mail: csjxie@nju.edu.cn
		\protect\\
	}
}

%
%

\markboth{Journal of \LaTeX\ Class Files,~Vol.~14, No.~8, August~2015}%
{Shell \MakeLowercase{\textit{et al.}}: Bare Demo of IEEEtran.cls for Computer Society Journals}
%



\IEEEtitleabstractindextext{%
\justify
\begin{abstract}
		In this paper, we propose a novel 3D registration paradigm,  \textit{Generative Point Cloud Registration}, which bridges advanced 2D generative models with 3D matching tasks to enhance registration performance. Our key idea is to generate cross-view consistent image pairs that are well-aligned with the source and target point clouds, enabling geometry-color feature fusion to facilitate robust matching. To ensure high-quality matching, the generated image pair should feature both  \textit{2D-3D geometric consistency} and \textit{cross-view texture consistency}. 
		To this end, we introduce \textit{DepthMatch-ControlNet} and \textit{LiDARMatch-ControlNet}, two matching-specific, controllable 2D generative models. 
        Specifically, for depth camera-based 3D registration with point clouds derived from the depth maps, \textit{DepthMatch-ControlNet} leverages the depth-conditioned generation capabilities of ControlNet to synthesize perspective-view RGB images that are geometrically consistent with depth maps, ensuring accurate 2D–3D alignment. Additionally, by incorporating a coupled  conditional denoising scheme and coupled prompt guidance, it further promotes cross-view feature interaction, guiding texture consistency generation. 
		To address LiDAR-based 3D registration with point clouds captured by LiDAR sensors, \textit{LiDARMatch-ControlNet} extends this framework by conditioning on paired equirectangular range maps projected from 360$^\circ$ LiDAR point clouds, generating corresponding panoramic RGB images. 
Our generative 3D registration paradigm is general and can be seamlessly integrated into a wide range of existing registration methods to improve their performance. Extensive experiments on the 3DMatch and ScanNet datasets (for depth-camera settings), as well as the Dur360BEV dataset (for LiDAR settings), demonstrate the effectiveness of our approach. 
\end{abstract}

}

\maketitle

\IEEEdisplaynontitleabstractindextext

%
\IEEEpeerreviewmaketitle

\IEEEraisesectionheading{\section{Introduction}\label{intro}} 
\IEEEPARstart{P}{oint} cloud registration is a problem of finding the optimal rigid transformation, comprising a 3D rotation and a 3D translation, which aligns the source and target point clouds precisely. It plays an important role in various downstream computer vision applications, such as 3D reconstruction, LiDAR SLAM, and object localization. However,  real-world challenges like low overlap and noisy points still hinder its adoption in broader real-world scenarios. 

    		\begin{figure}[t]
		\centering
		\includegraphics[width=\columnwidth]{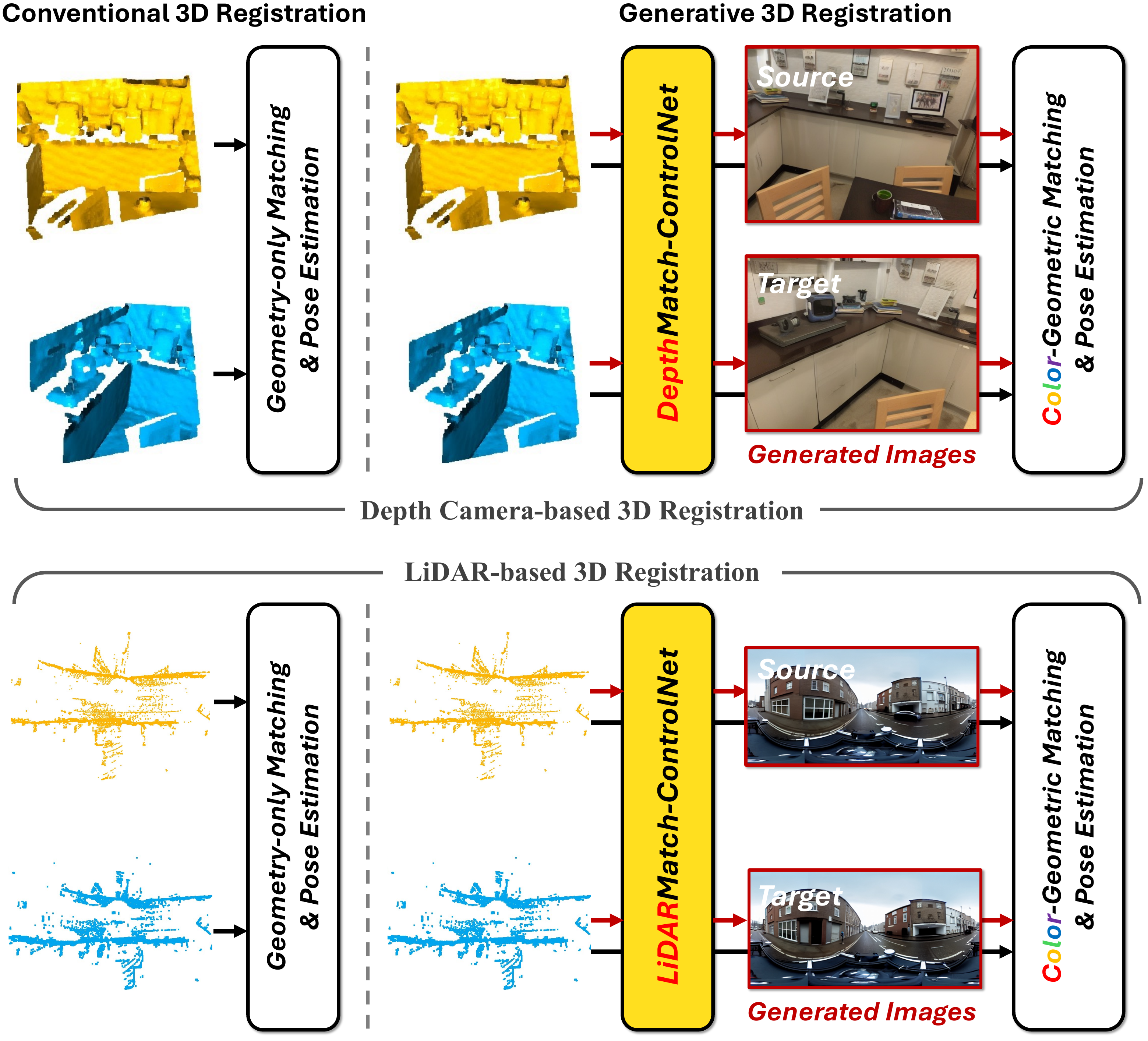}
        \vspace{-5mm}
		\caption{Paradigm comparison of our \textit{generative point cloud registration} with conventional methods. Unlike geometry-only matching in previous methods, our approach introduces \textit{Match-ControlNet}, a matching-specific 2D generative model that generates cross-view images pairs from point cloud data, providing rich color cues for enhanced geometric matching and pose estimation. In particular, we develop two variants: \textit{DepthMatch-ControlNet} and \textit{LiDARMatch-ControlNet} to handle depth camera-based and LiDAR-based point cloud registration tasks, respectively.}
		\label{highlight}
        \vspace{-5mm}
	\end{figure}
	
	Existing 3D registration methods can be roughly categorized into traditional approaches and data-driven deep  methods. The traditional approaches include optimization-based fine alignment methods~\cite{besl1992method,yang2013go}, which iteratively perform least-squares pose optimization for precise alignment, and handcrafted descriptor-based coarse alignment methods~\cite{rusu2008aligning,rusu2009fast}, which capture local geometry to establish correspondences for hypothesize-and-verify registration.
	Deep registration methods, whether end-to-end~\cite{yew2020rpm,yew2022regtr} or descriptor-based~\cite{huang2021predator,qin2022geometric}, exploit the power of deep neural networks to learn discriminative deep 3D features for robust matching. These deep methods significantly enhance the quality of estimated correspondences and improve registration accuracy. 
	
	Despite the impressive performance achieved by current point cloud registration methods, their robustness remains limited in challenging scenarios that contain low overlap, repetitive patterns, or noisy points. Recent RGB-D registration studies~\cite{yuan2023pointmbf,mu2024colorpcr} have shown that incorporating rich texture and semantic cues from RGB images would significantly enhance the distinctiveness of point cloud descriptors, leading to improved matching accuracy. However, in geometry-only point cloud registration, the RGB images corresponding to the point clouds are unavailable, and existing methods rely solely on 3D geometric information for correspondence estimation and pose calculation. 
	This raises an interesting question: ``\textit{Can we still leverage color information to enhance geometry-only point descriptors for enhanced 3D registration?}"

	Motivated by this question and inspired by the recent successes of generative AI models~\cite{ho2020denoising,song2020score,yang2023diffusion,li2025controlnet,rombach2022high,Zhang_2023_ICCV,wang2023freereg}, we introduce \textit{Generative Point Cloud Registration}, a new 3D matching paradigm that bridges the task gap between the 2D generative models and 3D matching tasks to enhance registration performance. 
	Our key idea is to generate cross-view image pairs that are well-aligned with the corresponding source and target point clouds. These images provide rich color information to complement geometric features, enabling more robust matching. 
	Unlike prevalent 2D generative models that focus on single image generation, our matching-specific image generation is pairwise. Importantly, to ensure high matching quality, the generated cross-view image pair should feature two key properties: \textit{2D-3D geometric consistency} and \textit{cross-view texture consistency}. 
	To achieve this, we introduce \textit{DepthMatch-ControlNet} and \textit{LiDARMatch-ControlNet}, two matching-specific, controllable 2D generative models as shown in Fig.~\ref{highlight}. 
	
	Specifically,  \textit{DepthMatch-ControlNet} is developed for the depth camera-based 3D registration, where the paired point clouds are derived from the depth maps. It leverages ControlNet's depth-conditioned generation capabilities~\cite{Zhang_2023_ICCV} to produce perspective-view RGB images geometrically aligned with depth maps, ensuring 2D-3D geometric consistency. Additionally, by incorporating coupled conditional denoising and coupled prompt guidance, it enables effective cross-view image feature interaction, achieving mutual texture message passing and thereby enhancing cross-view texture consistency. 
	It should be pointed out that this version can operate in both zero-shot and few-shot settings (with minimal fine-tuning samples), each providing valuable color information to enhance precision.
	\textit{LiDARMatch-ControlNet} is built for the LiDAR-based 3D registration that operates on the 360$^\circ$  LiDAR point clouds. It extends this framework by conditioning on paired equirectangular range maps projected from the LiDAR point clouds, and generates the corresponding panoramic RGB images with the geometric consistency and the cross-view texture consistency. 
		Finally, we propose a zero-shot geometric-color fusion mechanism that leverages pretrained large vision models (e.g., DINOv2~\cite{oquab2023dinov2} and Stable Diffusion~\cite{rombach2022high}) to extract discriminative zero-shot representations of generated images for enhancing geometric descriptors via weighted concatenation. 
 Moreover, our framework is general and can be integrated with various 3D registration methods to enhance their matching accuracy. Extensive experiments on the 3DMatch~\cite{zeng20173dmatch} and ScanNet~\cite{dai2017scannet} datasets (for depth-camera matching settings), and the Dur360BEV dataset~\cite{yuan2025dur360bev} (for LiDAR matching settings) validate the effectiveness of our proposed method.
	
	A preliminary version of this work was accepted by ICML'2025, and we  expand upon it with comprehensive discussions and significant improvements.  
	Specifically, our original manuscript focused solely on depth camera-based 3D registration using  proposed \textit{DepthMatch-ControlNet}. Beyond that, we now extend our \textit{Generative Point Cloud Registration} framework to handle LiDAR-based 3D registration with point clouds captured by LiDAR sensors. To this end, we develop a novel and effective \textit{LiDARMatch-ControlNet}. Here, we represent 360$^\circ$ LiDAR point clouds as equirectangular range maps, and innovatively take these range maps as condition to guide the generation of geometrically and texturally consistent panoramic images (see Sec.~\ref{outdoormatch}). To the best of our knowledge, this is the first successful exploration of LiDAR point cloud-to-panoramic image generation. 
Moreover, we provide a theoretical analysis of the coupled denoising diffusion mechanism used in both \textit{DepthMatch-ControlNet} and \textit{LiDARMatch-ControlNet}, showing that it effectively models the joint distribution of cross-view images and enables consistent generation (see Sec.~\ref{couplemodel}). Finally, we supplement our work with extensive experimental comparisons on Dur360BEV dataset, verifying the effectiveness of our newly introduced designs (see Sec.~\ref{comparequa}). 
    
	To summarize, our main contributions are as follows:
	\begin{itemize}
		\item We propose a new \textit{Generative Point Cloud Registration} paradigm, aimed at generating cross-view image pairs for both source and target point clouds, thereby providing rich color information for effective geometric-color feature fusion and improved matching quality. 
        \vspace{1mm}
		\item For depth camera-based 3D registration, we develop an effective \textit{DepthMatch-ControlNet} for matching-specific, pairwise perspective-view image generation. It incorporates depth-conditioned generation, coupled conditional denoising, and coupled prompt guidance to ensure that the generated image pairs maintain {2D-3D geometric consistency} and {cross-view texture consistency}.
        \vspace{1mm}
		\item For LiDAR-based 3D registration, we introduce a novel \textit{LiDARMatch-ControlNet}, where we innovatively represent the 360$^\circ$ LiDAR point clouds as the equirectangular range maps to guide the generation of geometrically and texturally consistent panoramic RGB images. To the best of our knowledge, this is the first successful realization of LiDAR point cloud-to-panoramic image generation.      
        \vspace{1mm}
		\item Our \textit{Generative Point Cloud Registration} framework is general and plug-and-play. Benefiting from our effective \textit{zero-shot geometric-color feature fusion} and \textit{XYZ-RGB fusion} schemes, it can be integrated with various 3D registration approaches to provide free-lunch color information, enhancing their registration  accuracy. 
	\end{itemize}	

        		\begin{figure*}[t]
		\centering
		\includegraphics[width=\textwidth]{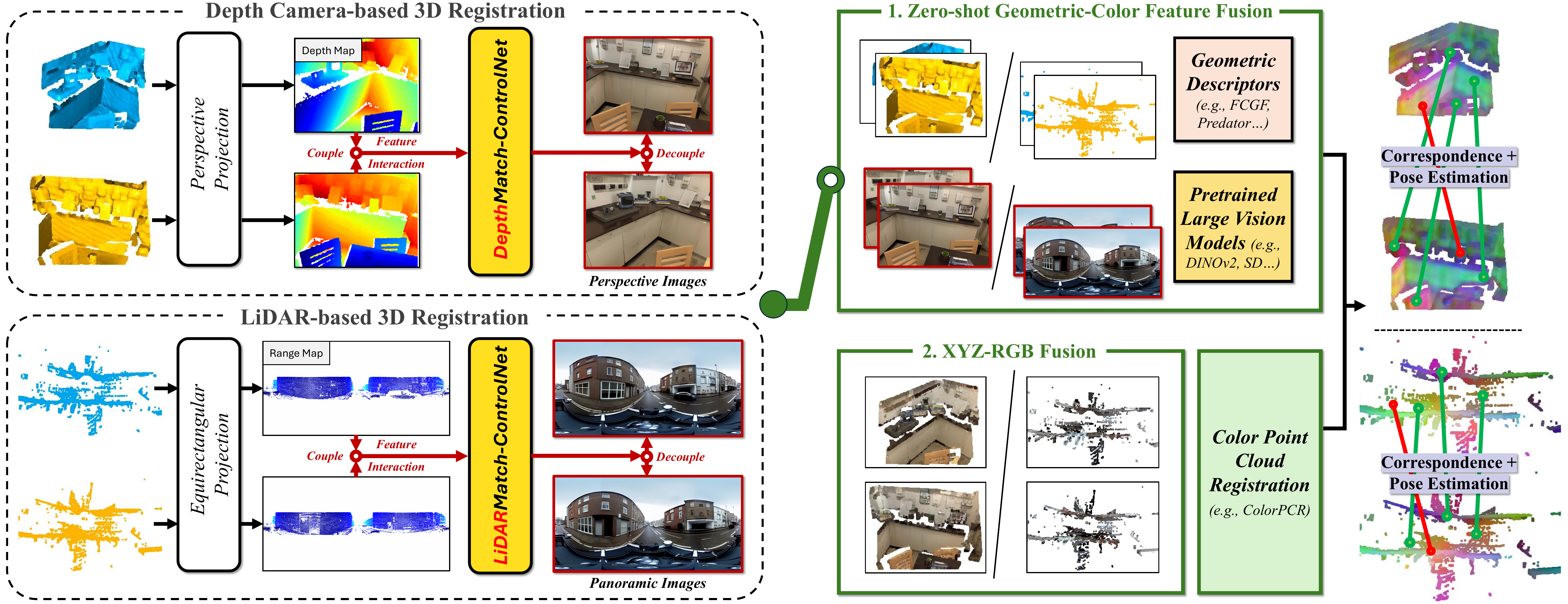}
		\caption{Pipeline of \textit{Generative Point Cloud Registration}. 
			Given a pair of source and target point clouds (captured by either a depth camera  or a LiDAR sensor), we first employ the corresponding \textit{DepthMatch-ControlNet} (\textit{LiDARMatch-ControlNet}) to synthesize perspective-view (panoramic) RGB images.
			Next, we employ either zero-shot geometric-color feature fusion or XYZ-RGB fusion to create color-enhanced geometric descriptors, enabling high-quality correspondence estimation and robust pose estimation.
		}
		\label{ftvis}
	\end{figure*}

	\section{Related Work}
	\label{related}
	\noindent\textbf{Traditional 3D Registration Methods.}  
	Traditional point cloud registration methods are typically categorized into coarse and fine registration approaches. Iterative Closest Point (ICP) \cite{besl1992method}, a prominent fine registration method, iteratively computes nearest-neighbor correspondences and performs least-squares optimization for pose estimation. Go-ICP \cite{yang2013go} enhances ICP's robustness to initialization errors through a branch-and-bound (BnB) global search. Trimmed ICP \cite{chetverikov2002trimmed} further improves robustness by optimizing over minimal subsets to handle outliers. Additional variants like \cite{sharp2002icp,fitzgibbon2003robust,bae2008method, gressin2013towards,deng2018ppfnet} also demonstrate promising precision in fine alignment. 
	Coarse registration methods generally combine handcrafted geometric descriptors with robust pose estimators, such as RANSAC. 
	\cite{johnson1999using} develops the spin image-based shape descriptors for surface matching and object recognition. 
	USC~\cite{tombari2010unique}  improves the feature descriptors using an shape context-aware unique local reference frame to improve matching accuracy. 
	SHOT~\cite{salti2014shot} introduces a 3D histogram-based feature using normal vectors to describe surface. 
	PFH \cite{rusu2008aligning} and FPFH \cite{rusu2009fast} constructs a discriminative and efficient local descriptor based on the oriented histogram with pairwise 3D representations. 
	Other notable coarse methods, including \cite{mohamad2014generalized, mohamad2015super, xu2019pairwise, huang2017v4pcs, ge2017automatic}, have also achieved impressive registration precision.
	
	\noindent\textbf{Learning-based Deep Registration Methods.}
	Deep registration methods primarily consist of end-to-end approaches and deep descriptor-based methods. 
	For end-to-end approaches, DCP~\cite{wang2019deep} introduces differentiable soft correspondences for SVD-based pose estimation. RPM-Net~\cite{yew2020rpm} incorporates the Sinkhorn layer and an annealing strategy to mitigate outlier inference. RegTR~\cite{yew2022regtr} designs an effective transformer-based correspondence regression module, addressing large-scale indoor scene registration in an end-to-end manner. 
	For deep descriptor-based methods, 3DMatch~\cite{zeng20173dmatch} employs a Siamese 3D CNN to extract local geometric features for patchwise matching. FCGF~\cite{choy2019fully} develops a fully convolutional network to learn dense 3D features for pointwise matching. Predator~\cite{huang2021predator} introduces a cross-attention transformer between point cloud pairs for overlap perception and robust registration. GeoTransformer~\cite{qin2022geometric} integrates geometric embeddings into the transformer, enhancing feature discrimination. RoITr~\cite{yu2023rotation} designs a rotation-invariant transformation to further improve the rotational robustness of geometric descriptors. 
	Other methods~\cite{wang2023zero,bai2020d3feat,li2022lepard,li2020iterative,choy2020deep,chen2023sira,fu2021robust,ao2023buffer} also demonstrate impressive performance in 3D registration. 
	Beyond traditional and learning-based frameworks, this work introduces a new paradigm:  \textit{Generative Point Cloud Registration}. By integrating advanced 2D generative models with the 3D registration domain, our approach generates complementary color information for input point cloud pairs, producing color-enhanced geometric descriptors to improve  matching precision. 

\section{Approach}\label{approach}
\subsection{Problem Formulation}
Given a pair of source and target point clouds $\mathcal{P} = \{\mathbf{p}_i \in \mathbb{R}^3 \mid i=1, \dots, N\}$ and $\mathcal{Q} = \{\mathbf{q}_i \in \mathbb{R}^3 \mid i=1, \dots, M\}$, point cloud registration seeks to recover their rigid transformation $\mathbf{T}=\{\mathbf{R}, \mathbf{t}\} \in SE(3)$, comprising a rotation $\mathbf{R} \in SO(3)$ and a translation $\mathbf{t} \in \mathbb{R}^3$, to align them precisely. The optimal rigid transformation is typically computed by solving:  
	\begin{equation}
		\min_{\mathbf{R}, \mathbf{t}} \sum_{(\mathbf{p}^*, \mathbf{q}^*) \in \mathcal{C}^*} \left\| \mathbf{R} \cdot \mathbf{p}^* + \mathbf{t} - \mathbf{q}^* \right\|_2^2,
	\end{equation}  
	where $\mathcal{C}^*$ denotes the ground-truth correspondences between source and target point clouds. However, $\mathcal{C}^*$ is generally unknown in practical usage, and we need estimate a set of putative correspondences through finding feature nearest neighbor among the pointwise geometric descriptors. 
    
\noindent\textbf{Depth Camera-based Point Cloud Registration.}
In our original manuscript, we primarily target the depth camera-based 3D registration where the point clouds are derived from the depth maps as in~\cite{zeng20173dmatch,dai2017scannet}. Due to the limited field-of-view, each point cloud typically captures only a partial observation of the scene. Thus, the resulting source and target point clouds (captured from different viewpoints) would suffer from partial overlap and incomplete geometry, increasing the challenge of accurate correspondence estimation and transformation recovery.

\noindent\textbf{LiDAR-based Point Cloud Registration.} 
Beyond the depth camera-based 3D registration describe above, our extended framework further generalizes the \textit{Generative Point Cloud Registration} paradigm to LiDAR-based 3D registration with 360$^\circ$ point clouds captured by LiDAR sensors.  
In this setting, two temporally adjacent LiDAR sweeps are treated as the source and target point clouds for geometric matching. Due to the sparsity of LiDAR data, robust LiDAR matching remains a challenging problem.

\subsection{Motivation} \label{motivate}
	Recent RGB-D point cloud registration methods~\cite{yuan2023pointmbf, mu2024colorpcr} have demonstrated that RGB images can significantly enhance geometry-only descriptors by providing rich color and semantic information. This enhancement facilitates the construction of higher-quality correspondences, leading to more robust registration.
	
	However, in the context of 3D matching tasks, we focus on geometry-only registration using pure point clouds, as the relevant RGB data is unavailable. To overcome this limitation, we introduce \textit{Generative Point Cloud Registration}, a general framework designed to generate high-quality RGB data for both point clouds, enabling geometric-color feature fusion for enhanced matching. Notably, unlike the conventional single-image generation focused by prevalent 2D generative models~\cite{rombach2022high,Zhang_2023_ICCV}, our matching-specific image generation is pairwise, which should satisfy two key criteria: 
	
	\noindent\textbf{(i) 2D-3D Geometric Consistency:}
	The generated images should preserve the geometric structure and spatial layout of their respective point clouds to ensure accurate pixel-to-point correspondences and avoid introducing noise; 
	
	\noindent\textbf{(ii) Cross-view Texture Consistency:}  
	The generated image pair should maintain consistent textures for correspondences. Otherwise, inconsistent textures would  reduce  feature similarity of correspondences, leading to mismatches. 

Given the significant disparities between depth camera-based point clouds and LiDAR-based point clouds (single-view vs. omnidirectional), as well as the distinct characteristics of their associated RGB images (perspective-view vs. panoramic), employing a unified point cloud-to-image generative model for both depth camera-based and LiDAR-based 3D registrations is inherently challenging. As such, we propose two variants: \textit{DepthMatch-ControlNet} and \textit{LiDARMatch-ControlNet}, designed to generate perspective-view RGB images for the depth-camera matching (see Sec.~\ref{indoormatch}) and panoramic RGB images for the LiDAR matching (see Sec.~\ref{outdoormatch}), respectively.

	\subsection{DepthMatch-ControlNet} \label{indoormatch}
In this section, we first introduce the \textit{DepthMatch-ControlNet}, which takes point cloud pairs derived from depth maps as input and generates the corresponding source and target perspective-view RGB images that exhibit strong 2D-3D geometric consistency and cross-view texture consistency for RGB-enhanced geometric matching. 
    
    \subsubsection{Zero-Shot Geometric Consistency Generation} \label{geogen}
	We first address the 2D-3D geometric consistency generation through  ControlNet~\cite{Zhang_2023_ICCV}, a variant of Stable Diffusion~\cite{rombach2022high} with spatially localized image conditions (e.g. Canny edge maps and depth maps). 
	
	\noindent\textbf{Stable Diffusion.} Stable Diffusion is a widely used latent diffusion model for text-to-image generation. It operates within the latent space of a pretrained autoencoder, where a denoiser $\boldsymbol{\epsilon}_\theta(\mathbf{x}_t ; t, \mathbf{c})$ (conditioned on the timestamp $t$ and tokenized text prompt $\mathbf{c}$) gradually refines the noisy latent feature $\mathbf{x}_t$ to clean one for image decoding. The denoiser follows a UNet architecture with an encoder, middle block, and skip-connected decoder, incorporating stacked transformer and residual modules. Each transformer module utilizes intra-image self-attention for contextual understanding and prompt-to-image cross-attention to guide generation.
	
	\noindent\textbf{ControlNet-driven 2D-3D Geometric Consistency.} ControlNet further equips the denoiser of Stable Diffusion with  a learnable encoder copy for encoding the conditional image $\mathbf{c}_{I}$, forming a conditional denoiser: $\tilde{\boldsymbol{\epsilon}}_\theta(\mathbf{x}_t ; t, \mathbf{c},  \mathbf{c}_{I})$.  
	Consequently, the encoded condition features is concatenated with the original encodings of  noisy latent representations $\mathbf{x}_t$ for conditional feature decoding via skip connections. Notably, ControlNet allows the use of depth maps as conditional inputs to generate RGB images that preserve geometric structures well-aligned with the provided depth prior.  This capability perfectly aligns with our objective and motivates us to convert the source and target point clouds into their corresponding depth maps, $\mathbf{D}_{\mathcal{P}}$ and $\mathbf{D}_{\mathcal{Q}} \in \mathbb{R}^{H \times W \times 1}$, via the intrinsic matrix. Then, each produced depth map can be independently used to condition ControlNet to produce the geometrically consistent image. 
	
				\begin{figure}[t]
		\centering
		\includegraphics[width=\columnwidth]{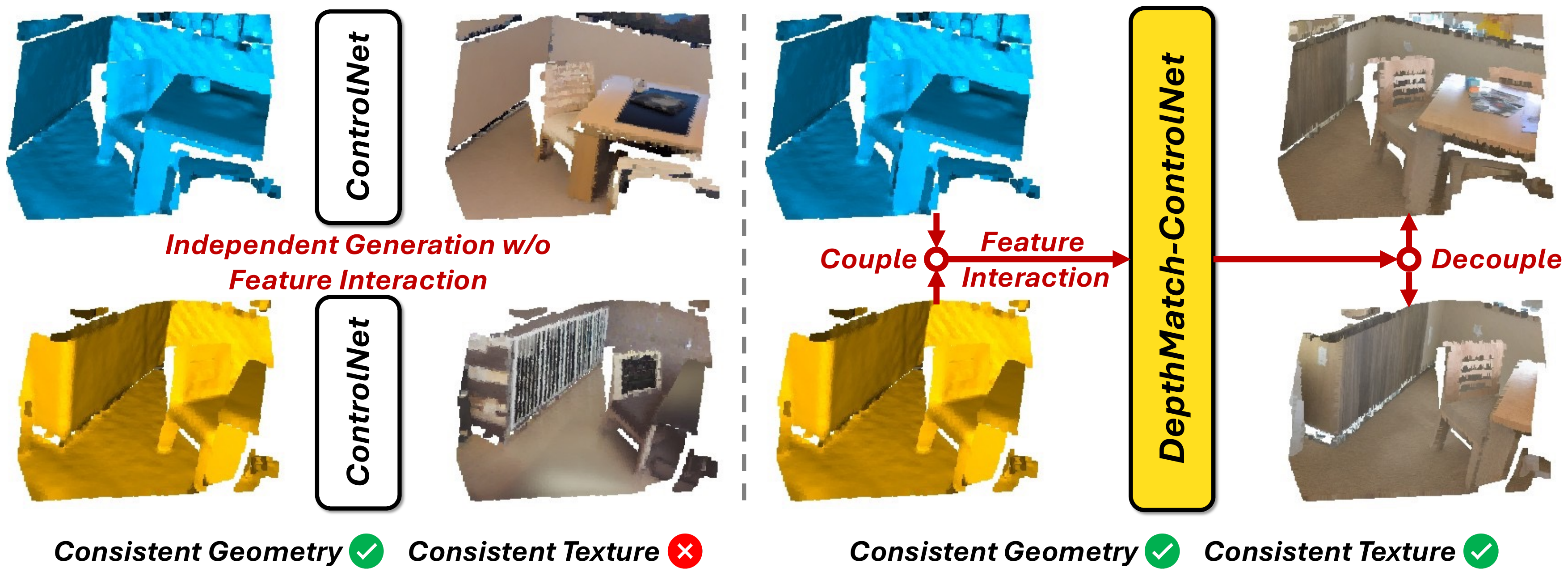}
		\vspace{-6mm}
		\caption{Instead of  independently performing ControlNet to generate source and target images, our \textit{DepthMatch-ControlNet} integrates their denoising diffusion generation processes into a unified framework, facilitating feature interaction (i.e., mutual texture message passing) and enhancing their cross-view texture consistency.
		}
		\label{matchcontrolnet}
		\vspace{-3mm}
	\end{figure}

	\subsubsection{Zero-Shot Texture Consistency Generation} \label{zeroconsis}
	Although the original ControlNet can generate source and target image pairs that are geometrically well-aligned with the given point cloud pair, the texture details of the corresponding regions between the generated image pair often differ as shown in Fig.~\ref{matchcontrolnet} (\textit{left}). This texture inconsistency primarily arises because the denoising processes for the source and target images operate independently as follows: 
	\begin{equation}
		\begin{split}
			\tilde{\boldsymbol{\epsilon}}_\theta(\mathbf{x}_t^{\mathcal{P}} ; t, \mathbf{c}, \mathbf{d}_{\mathcal{P}})\rightarrow \mathbf{x}_{t-1}^{\mathcal{P}}, \\
			\tilde{\boldsymbol{\epsilon}}_\theta(\mathbf{x}_t^{\mathcal{Q}} ; t, \mathbf{c}, \mathbf{d}_{\mathcal{Q}})\rightarrow \mathbf{x}_{t-1}^{\mathcal{Q}}, 
		\end{split}
	\end{equation}
	with each unaware of the colors produced by the other. 
	Here, $\mathbf{x}_t^\mathcal{P}, \mathbf{x}_t^\mathcal{Q}\in\mathbb{R}^{H'\times W' \times d}$ denote the noisy latent representations corresponding to source and target images; $\mathbf{d}_\mathcal{P}, \mathbf{d}_\mathcal{Q}\in\mathbb{R}^{H'\times W' \times d}$ represent the encoded features of depth maps $\mathbf{D}_\mathcal{P}$ and $\mathbf{D}_\mathcal{Q}$ via optimized zero convolutions of ControlNet.  
	This insight motivates us to combine source and target image denoising generation processes into a joint denoising pass, thereby enabling mutual texture message passing and promoting texture consistency (see Fig.~\ref{matchcontrolnet}). 
	
	Based on this motivation, we establish \textit{DepthMatch-ControlNet}, an improved ControlNet variant for matching-specific, conditional perspective-view image generation. Following Sec.~\ref{geogen}, we still take depth maps derived by point clouds as conditional images so as to inherit ControlNet's 2D-3D geometric consistency generation capability. 
	Additionally, we introduce two key designs: coupled conditional denoising and coupled prompt guidance to achieve the cross-view texture consistency generation. The details of these two designs are as follows:
	
	\noindent\textbf{Coupled Conditional Denoising.} To achieve mutual texture message passing, a straightforward approach is to build two denoisers and incorporate an additional cross-denoiser attention module to facilitate their message passing. However, running two denoisers simultaneously is inefficient, and such significant architectural changes would require extensive model fine-tuning.
	
	To enable effective cross-view message passing without any finetuning (i.e., zero-shot), we propose an efficient coupled conditional denoising scheme for joint, interactive source and target image generations. Specifically, we expand the noisy latent representation $\mathbf{x}_t^{\mathcal{P}(\mathcal{Q})}$ with shape $[H', W', d]$ to a coupled one $\mathbf{x}_t^{\mathcal{PQ}}$ with the extended shape $[2H', W', d]$. 
	Also, we vertically concatenate the source and target depth maps into a coupled one $\mathbf{D}_{\mathcal{PQ}}\in\mathbb{R}^{2H \times W \times 1}$, and further employ the ControlNet's zero convolutions to encode it into the condition features $\mathbf{d}_{\mathcal{PQ}} \in \mathbb{R}^{2H' \times W' \times d}$. 	 
	Consequently, without any architectural modifications or parameter fine-tuning, the original conditional denoiser can be directly employed for our coupled conditional denoising: 
	\begin{equation}\label{coupcond}
		\tilde{\boldsymbol{\epsilon}}_\theta\left(\mathbf{x}_t^{\mathcal{PQ}} ; t, \mathbf{c}, \mathbf{d}_{\mathcal{PQ}}\right)\rightarrow \mathbf{x}_{t-1}^{\mathcal{PQ}},
	\end{equation}
	forming the denoising Markov chain: $\mathbf{x}_T^{\mathcal{PQ}} \rightarrow \cdots \mathbf{x}_1^{\mathcal{PQ}} \rightarrow \mathbf{x}_0^{\mathcal{PQ}}$. Here, the initial coupled latent representation $\mathbf{x}_T^{\mathcal{PQ}}$ is sampled from a standard Gaussian distribution  $\mathcal{N}(\mathbf{0}, \mathbf{I})$. 
	To further clarify the cross-view texture message passing during our coupled denoising process, we formulate the self-attention mechanism within the denoiser as $SA(\mathbf{x}_t^{\mathcal{PQ}}) =$
	\begin{equation}\label{sa}
		\begin{split}
			\operatorname{softmax}(\frac{(\mathbf{x}_t^{\mathcal{PQ}}\mathbf{W}^q)(\mathbf{x}_t^{\mathcal{PQ}}\mathbf{W}^k)^\top}{\sqrt{d}})(\mathbf{x}_t^{\mathcal{PQ}}\mathbf{W}^v),
		\end{split}
	\end{equation}
	where $\mathbf{W}^q$, $\mathbf{W}^k$ and $\mathbf{W}^v\in\mathbb{R}^{d\times d}$ denote the projection matrices for queries, keys and values, respectively. 
	Eq.~\ref{sa} illustrates that by coupling the source and target noisy latent representations, each feature element can establish long-range dependencies with all feature elements from both the source and target feature maps, allowing effective cross-view feature interaction and texture-aware message passing for promoting texture consistency generation.
	
	\noindent\textbf{Coupled Prompt Guidance.} Although the aforementioned coupled denoising generative mechanism has provided the essential components for  texture consistency generation, the denoiser still fails to produce the image pair with expected cross-view texture consistency. The core reason is that the denoiser does not know what kind of image the user expects to generate, and we need to tell it what to do. 
	Our important finding is that when we use a specific prompt (named \textit{coupled prompt}) as below to tell our \textit{DepthMatch-ControlNet} to produce the vertically stacked images with consistent  layout and elements:	
	``\textcolor{black}{{\textit{Generate two vertically stacked images that are captured from the different viewpoints in a same scene. The images should feature the same environment—whether indoor or outdoor, like a living room, office, street, or natural landscape—with very subtle differences between them. Overall, the layout and key elements remain the same.}}}",  
	the denoiser can naturally be guided to recover consistent textures without any model fine-tuning.  To the best of our knowledge, we are the first to uncover and utilize this inherent capability of the pre-trained ControlNet for zero-shot pairwise image generation enjoying both 2D-3D geometric consistency and cross-view texture consistency.

	\begin{figure}[t]
		\centering
		\includegraphics[width=\columnwidth]{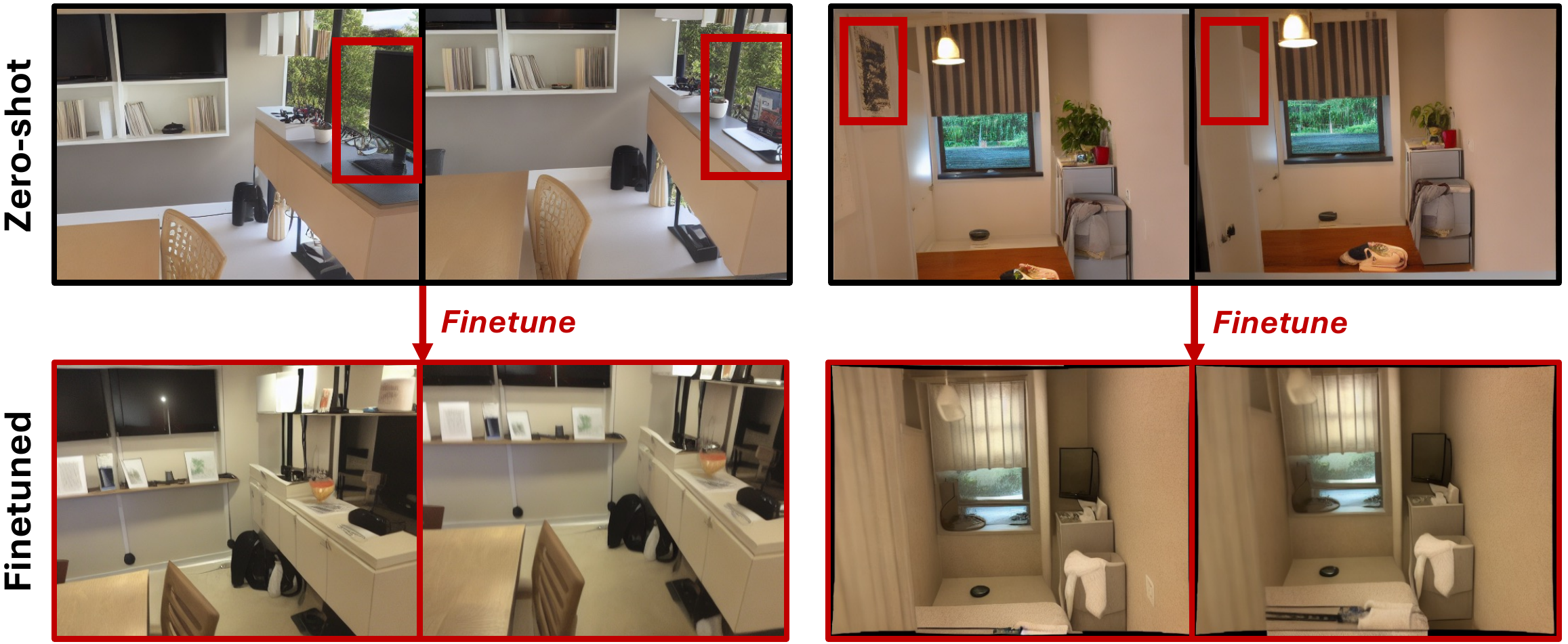}
		\caption{Compared to the \textit{zero-shot DepthMatch-ControlNet} (top), the \textit{finetuned DepthMatch-ControlNet} can tend to achieve higher 2D-3D geometric consistency and the cross-view texture consistency. 
		}
		\label{ftvis}
	\end{figure}
	
	\subsubsection{Few-Shot Consistency Fine-tuning} \label{fintune1}
	Although our zero-shot \textit{DepthMatch-ControlNet} above has demonstrated promising pairwise consistency generation capabilities,  Fig.~\ref{ftvis} shows that some corresponding regions may still exhibit geometric or texture inconsistency issues. To mitigate it, we further propose a few-shot finetuning mechanism to improve the consistency generation quality of our  \textit{DepthMatch-ControlNet}. 
	It's noted that we only finetune the  learnable encoder copy of \textit{DepthMatch-ControlNet} rather than the all parameters to preserve the powerful generation ability of Stable Diffusion. 
	Specifically, we first collect a set of RGB-depth pairs, $\{((\mathbf{I}_{\mathcal{P}}, \mathbf{D}_{\mathcal{P}}), (\mathbf{I}_{\mathcal{Q}}, \mathbf{D}_{\mathcal{Q}}))_j\}$ ($j$ denotes the sample index), as the training data. For each sample pair, we then concatenate its depth maps and RGB images into coupled ones $\{(\mathbf{I}_{\mathcal{P}\mathcal{Q}}, \mathbf{D}_{\mathcal{P}\mathcal{Q}})_j\}$. Finally, we use the loss function below to finetune the denoiser: 
	\begin{equation}\label{loss1}
		\mathcal{L} = \mathbb{E}_{\mathbf{x}_t^{\mathcal{PQ}}, t, \tilde{\mathbf{c}}, \mathbf{d}_{\mathcal{PQ}}, \boldsymbol{\epsilon}} \left[ \left\| \boldsymbol{\epsilon} - \tilde{\boldsymbol{\epsilon}}_\theta(\mathbf{x}^{\mathcal{PQ}}_t, t, \tilde{\mathbf{c}}, \mathbf{d}_{\mathcal{PQ}}) \right\|_2^2 \right],
	\end{equation} 
	where the noise variable $\boldsymbol{\epsilon}\sim \mathcal{N}(\mathbf{0}, \mathbf{I})$ and $\mathbf{x}^{\mathcal{PQ}}_t$ represents the diffused latent representation of the coupled image $\mathbf{I}_{\mathcal{PQ}}$; $\tilde{\mathbf{c}}$ denotes the token sequence of our \textit{coupled prompt}. 
	Our experiments show that even with a limited number of samples ($\sim$3K), few-shot finetuning can effectively improve the quality of consistency generation.

\subsection{LiDARMatch-ControlNet} \label{outdoormatch}
In addition to generating perspective-view images conditioned on point cloud pairs derived from single-view depth maps in Sec.~\ref{indoormatch}, we further propose the \textit{LiDARMatch-ControlNet}, which takes as input a pair of 360-degree, temporally adjacent LiDAR point clouds and generates paired panoramic RGB images that preserve both geometric and texture consistency for RGB-enhanced LiDAR point cloud registration. \textit{Notably, to the best of our knowledge, there are no off-the-shelf models for LiDAR point cloud-to-panoramic image generation. We are the first to successfully realize this task, which requires careful design of various components, including the training data source, 2D/3D data representations, and network architecture}, as detailed below.

\subsubsection{Panoramic Image Representation}\label{imgrep}
To enable high-quality panoramic image generation from omnidirectional LiDAR point clouds, it is first necessary to determine an appropriate  panoramic image representation that effectively captures the complete spatial context of the surrounding scene. Unlike conventional perspective-view images, panoramic representations must preserve continuity across a wide field of view while remaining compatible with image generation models. 

Existing panoramic formats generally fall into two categories:
\textbf{(i) Multi-view perspective representations}, which divide the 360$^\circ$ scene into discrete, forward-facing subviews (e.g., front, left, right, rear). This representation is widely adopted in autonomous driving systems with multi-camera setups and is advantageous for preserving low-distortion views within each subregion~\cite{caesar2020nuscenes}. 
However, the imposed boundaries between adjacent subviews inherently disrupt the spatial continuity of the scene and segment the global context into disjoint fragments. Consequently, downstream models have to rely on explicit view-stitching strategies to fuse cross-view information, which significantly constrains their ability to capture whole geometric structures and consistent contextual dependencies across the entire scene. 
 \textbf{(ii) Unified panoramic representations}, e.g. equirectangular projections, unfold the entire spherical scene into a continuous 2D image domain. This format encodes the full horizontal (azimuth) field of view ($360^\circ$) and vertical (polar) extent ($180^\circ$), while preserving spatial continuity and maintaining geometric relationships across the entire scene. Such representations have been widely adopted in $360^\circ$ scene understanding tasks (e.g., panoramic semantic segmentation and virtual view synthesis) due to their compatibility with 2D convolutional networks and transformer-based vision backbones. 

In our framework, we adopt the equirectangular projection-based unified panoramic representation as the image format, due to its ability to preserve spatial continuity for global context modeling and its seamless architectural compatibility with diffusion-based generative backbones such as Stable Diffusion and ControlNet. 

\subsubsection{Panoramic LiDAR Point Cloud Representation} \label{pcrep}

Following the determination of the panoramic image format, we construct a corresponding panoramic geometric representation from the 360$^\circ$ LiDAR point clouds to serve as the spatial condition for our \textit{Panoramic Match-ControlNet}. Specifically, we project the raw LiDAR points into an equirectangular range map that is spatially aligned with the panoramic image domain, thus enabling fine-grained, pixel-wise conditioning for generative guidance.

Let us take the source LiDAR point cloud $\mathcal{P}$ as example (the target point cloud $\mathcal{Q}$ is the same), where each 3D point $\mathbf{p}_i = (x_i, y_i, z_i)$ is defined in the LiDAR sensor's coordinate system. We first transform each point into spherical coordinates $(\theta_i, \phi_i)$ as follows: 
\begin{equation}
\theta_i = \arccos({z_i}/{\|\mathbf{p}_i\|_2}), \quad \phi_i = \arctan2(y_i, x_i),
\end{equation}
where $\theta_i$ and $\phi_i$ denote the elevation and azimuth angles, respectively. We then adopt the equirectangular projection, which uniformly maps angular coordinates onto a $H \times W$ equirectangular grid, to discretize the spherical domain:
$u_i = \lfloor {W(\phi_i + \pi)}/{2\pi} \rfloor, v_i = \lfloor {H\theta_i}/{\pi} \rfloor$,
with $(u_i, v_i)$ indicating the pixel location corresponding to the point $\mathbf{p}_i$. The resulting equirectangular range map $\mathbf{D}_{\mathcal{P}}^{\text{equi}} \in \mathbb{R}^{H \times W \times 1}$  records the radial distance $\|\mathbf{p}_i\|_2$ at each pixel $(u_i, v_i)$. Similarly, the range map of the target point cloud is denoted as $\mathbf{D}_{\mathcal{Q}}^{\text{equi}}\in \mathbb{R}^{H \times W \times 1}$. 
In the presence of multiple points mapping to the same pixel, we retain the nearest point to ensure visibility consistency. 
This projection effectively flattens the irregular, omnidirectional LiDAR point cloud into an image-like structure aligned with the equirectangular RGB format above, facilitating range map-conditioned panoramic image generation.

\subsubsection{Panoramic Coupled Denoising Generation}
As performed in Sec.~\ref{indoormatch}, our \textit{LiDARMatch-ControlNet} still follows the coupled conditional denoising mechanism to generate  geometrically and texturally consistent panoramic image pairs from paired omnidirectional LiDAR point clouds. Specifically, given the source and target equirectangular range maps derived in Sec.~\ref{pcrep}, we vertically concatenate them into an unified conditional map $\mathbf{D}^{\text{equi}}_{\mathcal{PQ}}\in\mathbb{R}^{2H\times W\times 1}$. This map is then encoded via ControlNet's zero convolutions to produce conditional features $\mathbf{d}^{\text{equi}}_{\mathcal{PQ}}\in \mathbb{R}^{2H' \times W' \times d}$, which is used to condition the denoising process over the stacked latent panoramic representation $\tilde{\mathbf{x}}_t^{\mathcal{PQ}} \in \mathbb{R}^{2H' \times W' \times d}$. 
This coupling enables cross-view attention within the UNet, allowing mutual feature interaction across the source and target images without modifying the original model architecture:
\begin{equation}
\tilde{\boldsymbol{\epsilon}}_\theta\left(\tilde{\mathbf{x}}_t^{\mathcal{PQ}} ; t, \tilde{\mathbf{c}}, \mathbf{d}^{\text{equi}}_{\mathcal{PQ}}\right)\rightarrow \tilde{\mathbf{x}}_{t-1}^{\mathcal{PQ}}, 
\end{equation} 
forming the denoising Markov chain: $\tilde{\mathbf{x}}_T^{\mathcal{PQ}} \rightarrow \cdots \tilde{\mathbf{x}}_1^{\mathcal{PQ}} \rightarrow \tilde{\mathbf{x}}_0^{\mathcal{PQ}}$. The initial coupled latent representation $\tilde{\mathbf{x}}_T^{\mathcal{PQ}}$ is sampled from  $\mathcal{N}(\mathbf{0}, \mathbf{I})$. 
Here, $\tilde{\mathbf{c}}$ represents the token sequence of \textit{panoramic coupled prompt}: 
``\textcolor{black}{{\textit{Generate two vertically stacked surround-view panoramas that are captured from a self-driving vehicle. The panoramas should feature the same urban street environment, like wide roads, vehicles, buildings, and trees—with very subtle differences between them. Overall, the layout and key elements remain the same.}}}",  
that further promotes the model to generate vertically stacked panoramic images with consistent scene layout and subtle cross-view differences. However, since there is no off-the-shelf pretrained ControlNet conditioned on range maps for panoramic image generation, zero-shot generation, as achieved in the perspective setting (see Sec.~\ref{geogen} and Sec.~\ref{zeroconsis}), is not feasible. Instead, effective panoramic generation requires a consistency fine-tuning mechanism, which we detail in the following subsection. 

	\begin{table}[t]
	\centering
	\caption{Comparison of LiDAR datasets with respect to camera configurations and the azimuthal and polar field of view of panoramic images.}
	\vspace{-3mm}
	\resizebox{1\columnwidth}{!}{
		\begin{tabular}{lccc}
			\toprule[1.5pt]
			\textbf{Datasets}    & \textbf{Camera} &  \textbf{FoV Azm.} &  \textbf{FoV Plr.} \\
			\midrule[1.pt]
			KITTI~\cite{Geiger2012CVPR}    & 1 stereo cam & $<360^\circ$ &  $<180^\circ$   \\
			KITTI-360~\cite{Liao2022PAMI}    & 1 stereo + 2 fisheye  & $360^\circ$ & $120^\circ$  \\
			Waymo~\cite{sun2020scalability}    & 5 perspective   &  $<360^\circ$ &  $<180^\circ$  \\
			nuScenes~\cite{caesar2020nuscenes}    & 5 perspect. + 1 fisheye  &  $360^\circ$ & $40^\circ$  \\
			\midrule
			\textbf{Dur360BEV}~\cite{yuan2025dur360bev}    & 1 spherical cam & $\textbf{360}^\circ$ & $\textbf{180}^\circ$  \\
			\bottomrule[1.5pt]
	\end{tabular}}
	\label{lidarcomp}
\end{table}

\subsubsection{Panoramic Consistency Fine-tuning}
To achieve high-quality geometric and texture consistency in our \textit{LiDARMatch-ControlNet}, we similarly adopt a fine-tuning approach as described in Sec.~\ref{fintune1}. However, in the panoramic scenario, the selection of appropriate panoramic data for supervision becomes crucial. We have analyzed several widely-used LiDAR datasets in terms of camera setup and field of view (FoV), as summarized in Table~\ref{lidarcomp}.

Considering that our goal is to achieve comprehensive consistency across omnidirectional scenes, we select the Dur360BEV dataset~\cite{yuan2025dur360bev} for fine-tuning since it uniquely provides genuinely panoramic images captured by a spherical camera system. This dataset offers a complete 360$^\circ$ horizontal FoV (azimuth) and a 180° vertical FoV (polar), ensuring that the generated panoramic pairs are supervised by complete and coherent spatial information without any missing views or segmentation, which is crucial for maintaining global consistency. 
Specifically, we first collect a set of paired RGB panoramas and corresponding LiDAR-derived range maps, denoted as $\{((\mathbf{I}^{\text{equi}}_{\mathcal{P}}, \mathbf{D}^{\text{equi}}_{\mathcal{P}}), (\mathbf{I}^{\text{equi}}_{\mathcal{Q}}, \mathbf{D}^{\text{equi}}_{\mathcal{Q}}))_j\}$. We vertically concatenate each panorama and its corresponding range map into unified coupled formats $\{(\mathbf{I}^{\rm equi}_{\mathcal{PQ}}, \mathbf{D}^{\text{equi}}_{\mathcal{PQ}})_j\}$. Finally, we fine-tune only the learnable encoder of \textit{LiDARMatch-ControlNet} using the following loss function:
\begin{equation}\label{loss2}
\mathcal{L} = \mathbb{E}_{\tilde{\mathbf{x}}_t^{\mathcal{PQ}}, t, \tilde{\mathbf{c}}, \mathbf{d}^{\text{equi}}_{\mathcal{PQ}}, \boldsymbol{\epsilon}} \left[ \| \boldsymbol{\epsilon} - \tilde{\boldsymbol{\epsilon}}_\theta(\tilde{\mathbf{x}}^{\mathcal{PQ}}_t, t, \tilde{\mathbf{c}}, \mathbf{d}_{\mathcal{PQ}}^{\text{equi}}) \|_2^2 \right],
\end{equation}
where the noise $\boldsymbol{\epsilon}\sim \mathcal{N}(\mathbf{0}, \mathbf{I})$ and  $\tilde{\mathbf{x}}^{\mathcal{PQ}}_t$ is the diffused latent representation of the coupled panoramic images $\mathbf{I}^{\text{equi}}_{\mathcal{PQ}}$. Our experimental results indicate that limited fine-tuning samples (approximately 10K panoramic pairs from Dur360BEV~\cite{yuan2025dur360bev}) can achieve consistent panoramic image generation. Please refer to Sec.~\ref{comparequa} for more details on how these paired training samples are constructed.

\subsection{Theoretical Analysis of Coupled Denoising Model}\label{couplemodel}
In both the aforementioned \textit{DepthMatch-ControlNet} and \textit{LiDARMatch-ControlNet}, we have introduced our proposed coupled conditional denoising mechanism in generating cross-view consistent perspective-view and panoramic images. In this section, we provide a theoretical interpretation of how the coupled denoising mechanism effectively models the joint distribution of cross-view consistent images, thereby enabling high-quality consistent generation.

Specifically, let $(\mathbf{x}^\mathcal{P}, \mathbf{x}^\mathcal{Q}, \mathbf{d}_\mathcal{P}, \mathbf{d}_{\mathcal{Q}})\sim p_{data}$ denote the paired clean cross-view perspective (or panoramic) images and depth (or range) map conditions in the latent space ($p_{data}$ represents the sample distribution). The likelihood over the cross-view image pair can be expressed as $\mathbb{E}_{p_{data}}[p_\theta(\mathbf{x}^{\mathcal{P}}, \mathbf{x}^\mathcal{Q}| \mathbf{d}_\mathcal{P}, \mathbf{d}_\mathcal{Q})]$. 
By concatenating the cross-view images and condition maps into unified variables $\mathbf{x}^\mathcal{PQ} = \operatorname{Cat}([\mathbf{x}^\mathcal{P}; \mathbf{x}^\mathcal{Q}])$ and $\mathbf{d}_\mathcal{PQ} = \operatorname{Cat}([\mathbf{d}_\mathcal{P}; \mathbf{d}_\mathcal{Q}])$, we can model the joint distribution of cross-view images via the likelihood over the concatenated representation: $\mathbb{E}_{p_{data}}[p_\theta(\mathbf{x}^\mathcal{PQ}| \mathbf{d}_\mathcal{PQ})]$. 

The diffusion process progressively introduces the Gaussian noise into this clean image pair, which gradually converts them into the noise distribution $\mathbf{x}_T^\mathcal{PQ}\sim \mathcal{N}(\mathbf{0}, \mathbf{I})$, forming a Markov chain $\mathbf{x}^\mathcal{PQ}=\mathbf{x}_0^\mathcal{PQ}\rightarrow \mathbf{x}_1^\mathcal{PQ} \cdots \rightarrow \mathbf{x}_T^\mathcal{PQ}$ with  $\mathbf{x}_t^{\mathcal{PQ}}\sim \mathcal{N}(\mathbf{x}_t^{\mathcal{PQ}}; \sqrt{\bar{\alpha}_t}\mathbf{x}_0^{\mathcal{PQ}}, (1-\bar{\alpha}_t)\mathbf{I})$.

Then, the reverse process aims to learn  a denoising network, a parameterized normal distribution $p_\theta(\mathbf{x}^\mathcal{PQ}_{t-1}|\mathbf{x}^\mathcal{PQ}_t, \mathbf{d}_\mathcal{PQ}):=\mathcal{N}(\mathbf{x}^\mathcal{PQ}_{t-1} ; \boldsymbol{\mu}_\theta(\mathbf{x}_t^\mathcal{PQ}, t, \mathbf{d}_\mathcal{PQ}), \beta_{t}\mathbf{I})$ (prompt term is removed for simplicity), to progressively denoise the noisy data $\mathbf{x}_T^\mathcal{PQ}$ into the clean one $\mathbf{x}_0^\mathcal{PQ}$, forming a reverse Markov chain $\mathbf{x}_T^\mathcal{PQ}\rightarrow \mathbf{x}_{T-1}^\mathcal{PQ}\rightarrow \cdots \rightarrow \mathbf{x}_0^\mathcal{PQ}$. Here, $\boldsymbol{\mu}_\theta(\mathbf{x}^\mathcal{PQ}_t, t, \mathbf{d}_{\mathcal{PQ}})$ indicates the parameterized mean of the normal distribution. 
Following~\cite{ho2020denoising}, the evidence lower bound of the log likelihood over the training data can be derived as the optimization objective for training the denoising network: $\mathcal{L}_{\text{ELBO}}=$
\begin{equation}\small
\begin{split}
&\mathbb{E}_{q}\Big[{\log p_\theta(\mathbf{x}^{\mathcal{PQ}}_0 \mid \mathbf{x}^{\mathcal{PQ}}_1, \mathbf{d}_{\mathcal{PQ}})}-\operatorname{D}_{\rm KL}(q(\mathbf{x}_{T}^\mathcal{PQ}|\mathbf{x}_{0}^\mathcal{PQ})|| p(\mathbf{x}_{T}^\mathcal{PQ})) \\
&-\sum_{t>1}{\operatorname{D_{KL}}(q(\mathbf{x}^{\mathcal{PQ}}_{t-1} \mid \mathbf{x}^{\mathcal{PQ}}_t, \mathbf{x}^{\mathcal{PQ}}_0) || p_\theta(\mathbf{x}^{\mathcal{PQ}}_{t-1} \mid \mathbf{x}^{\mathcal{PQ}}_t, \mathbf{d}_{\mathcal{PQ}}))}. 
\end{split}
\end{equation}
Here, the third term is the core loss function, referred to as the denoising matching loss. Through the reparameterization trick \(\mathbf{x}_0^{\mathcal{PQ}} = \frac{\mathbf{x}_t^{\mathcal{PQ}} - \sqrt{1 - \bar{\alpha}_t} \boldsymbol{\epsilon}}{\sqrt{\bar{\alpha}_t}}\) \(\left(\boldsymbol{\epsilon} \sim \mathcal{N}(\mathbf{0}, \mathbf{I})\right)\), the mean of the posterior distribution \(q(\mathbf{x}_{t-1}^\mathcal{PQ} | \mathbf{x}_t^\mathcal{PQ}, \mathbf{x}_0^\mathcal{PQ})\) can be rewritten as $\boldsymbol{\mu}_q(\mathbf{x}_t^{\mathcal{PQ}}, \mathbf{x}_0^{\mathcal{PQ}}) = \frac{1}{\sqrt{\alpha_t}} \mathbf{x}_t^{\mathcal{PQ}} - \frac{1 - \alpha_t}{\sqrt{1 - \bar{\alpha}_t} \sqrt{\alpha_t}} \boldsymbol{\epsilon}$.
Thus, the mean of the denoising network (i.e., prior distribution) can be modeled as $\boldsymbol{\mu}_\theta(\mathbf{x}_t^{\mathcal{PQ}}, t, \mathbf{d}_{\mathcal{PQ}}) :=$
\begin{equation}\small
\begin{split}
 \frac{1}{\sqrt{\alpha_t}} \mathbf{x}_t^{\mathcal{PQ}} - \frac{1 - \alpha_t}{\sqrt{1 - \bar{\alpha}_t} \sqrt{\alpha_t}} \boldsymbol{\epsilon}_\theta(\mathbf{x}_t^{\mathcal{PQ}}, t, \mathbf{d}_{\mathcal{PQ}}).
\end{split}
\end{equation}
Consequently, minimizing the KL divergence between the posterior distribution and the prior distribution in the denoising matching loss can be converted to optimizing the parameters of the noise network \(\boldsymbol{\epsilon}_\theta(\mathbf{x}_t^{\mathcal{PQ}}, t, \mathbf{d}_{\mathcal{PQ}})\) to approximate the noise \(\boldsymbol{\epsilon}\). The final loss function can be formulated as:
\begin{equation}\label{loss3}
    \mathcal{L} = \mathbb{E}_{\mathbf{x}_t^{\mathcal{PQ}}, t, \mathbf{d}_{\mathcal{PQ}}, \boldsymbol{\epsilon} \sim \mathcal{N}(\mathbf{0}, \mathbf{I})} \left[ \left\| \boldsymbol{\epsilon} - \boldsymbol{\epsilon}_\theta(\mathbf{x}_t^{\mathcal{PQ}}, t, \mathbf{d}_{\mathcal{PQ}}) \right\|_2^2 \right],
\end{equation}
which is equivalent to our practical denoising losses as in Eq.~\ref{loss1} and Eq.~\ref{loss2} (the prompt term is omitted for simplicity). This indicates that our coupled denoising mechanism implicitly models the joint model of the cross-view images for consistency generation. 
In particular, since the cross-view images inherently exhibit consistency, maximizing the likelihood of their joint distribution through the loss function Eq.~\ref{loss3} above can effectively promote our learned coupled denoising network to generate geometrically and texturally consistent perspective-view/panoramic RGB images, thereby improving the matching quality. 
Please refer to Appendix A for more details.

	\begin{table*}[h]
	\centering
	\caption{Comparison of the methods on rotation, translation, and Chamfer distance on \textbf{ScanNet}~\cite{dai2017scannet} benchmark dataset.}
	\vspace{-3mm}
	\resizebox{1\textwidth}{!}{
		\begin{tabular}{lccccccccccccccc}
			\toprule[1.8pt]
			&   \multicolumn{5}{c}{\textbf{Rotation} (\textit{deg})} & \multicolumn{5}{c}{\textbf{Translation} (\textit{cm})} & \multicolumn{5}{c}{\textbf{Chamfer} (\textit{mm})} \\
			\cmidrule(lr){2-16} 
			&   \multicolumn{3}{c}{\textbf{Accuracy} $\uparrow$} & \multicolumn{2}{c}{\textbf{Error}$\downarrow$} &  \multicolumn{3}{c}{\textbf{Accuracy} $\uparrow$} & \multicolumn{2}{c}{\textbf{Error}$\downarrow$} & \multicolumn{3}{c}{\textbf{Accuracy} $\uparrow$} & \multicolumn{2}{c}{\textbf{Error}$\downarrow$} \\
			\textbf{Methods}  & 5 & 10 & 45 & Mean & Med. & 5 & 10 & 25 & Mean & Med. & 1 & 5 & 10 & Mean & Med. \\
			\midrule[1.2pt]
			FPFH~\cite{rusu2009fast}  & 41.4 & 56.7 & 73.3 & 39.2 &  7.1 & 17.5 & 35.1 & 50.9 & 79.5 & 23.5 & 32.3 & 48.0 & 53.0 & 159.6 &  6.5  \\ 
			Lepard~\cite{li2022lepard}  & 63.3 & 75.5 & 84.1 & 24.9 &  3.3 & 31.3 & 56.4 & 72.3 & 48.4 &  8.1 & 51.6 & 69.1 & 73.0 & 89.2 &  0.9  \\
			RegTR~\cite{yew2022regtr}  &  72.5 & 83.8 & 94.1 & 10.2 &  2.3 & 44.3 & 65.6 & 80.0 & 27.7 &  5.8 & 61.0 & 76.6 & 80.9 & 54.0 &  0.5 \\
			RoITr~\cite{yu2023rotation} & 70.0 & 77.5 & 83.7 & 24.1 &  2.3 & 40.3 & 62.3 & 75.1 & 45.6 &  6.5 & 58.8 & 72.4 & 75.4 & 94.1 &  0.6 \\ 
			\midrule[1.2pt]
			FCGF~\cite{choy2019fully} &  78.9 & 84.2 & 87.5 & 19.4 &  1.5 & 55.3 & 70.7 & 79.7 & 37.8 &  4.3 & 67.3 & 78.2 & 80.3 & 100.7 &  0.4  \\
			Generative FCGF$^{\rm DINO_{v2}}$ & 81.0 & 86.2 & 89.4 & 16.5 &  1.4 & \textbf{57.3} & 72.6 & 80.9 & 33.9 &  \textbf{4.0} & \textbf{68.9} & 79.5 & 81.5 & 97.4 &  \textbf{0.3} \\
			Generative FCGF$^{\rm SD}$ & \textbf{82.9} & \textbf{90.0} & \textbf{94.4} &  \textbf{8.4} &  \textbf{1.6} & 56.4 & \textbf{73.0} & \textbf{82.7} & \textbf{21.7} &  4.1 & 67.7 & \textbf{80.9} & \textbf{83.7} & \textbf{66.0} &  0.4  \\ 
			\textit{Improvement} $\uparrow$ & \textcolor{teal}{\textit{4.0}}& \textcolor{teal}{\textit{5.8}}& \textcolor{teal}{\textit{6.9}}&\textcolor{teal}{\textit{11.0}} & \textcolor{teal}{\textit{0.1}}& \textcolor{teal}{\textit{2.0}}& \textcolor{teal}{\textit{2.3}}& \textcolor{teal}{\textit{3.0}}& \textcolor{teal}{\textit{16.1}}& \textcolor{teal}{\textit{0.3}}& \textcolor{teal}{\textit{1.6}}& \textcolor{teal}{\textit{2.7}}& \textcolor{teal}{\textit{3.4}} & \textcolor{teal}{\textit{34.7}} & \textcolor{teal}{\textit{0.1}}\\
			\midrule[1.2pt]
			Predator~\cite{huang2021predator}  & 64.3 & 75.2 & 82.6 & 26.3 &  3.2 & 30.1 & 54.8 & 69.2 & 48.7 &  8.4 & 50.8 & 66.9 & 70.6 & 93.2 &  1.0  \\
			Generative Predator$^{\rm DINO_{v2}}$ & 67.0 & 78.0 & 87.2 & 19.0 &  3.0 & 30.7 & 56.0 & 70.3 & 41.4 &  8.1 & 52.0 & 67.8 & 71.3 & 79.2 &  0.9  \\
			Generative Predator$^{\rm SD}$ &\textbf{70.7} & \textbf{81.3} & \textbf{88.7} & \textbf{17.0} &  \textbf{2.8} & \textbf{33.0} & \textbf{59.4} & \textbf{73.3} & \textbf{36.6} &  \textbf{7.5} & \textbf{54.7} & \textbf{70.8} & \textbf{74.2} & \textbf{72.5} &  \textbf{0.8}  \\
			\textit{Improvement} $\uparrow$ & \textcolor{teal}{\textit{6.4}}& \textcolor{teal}{\textit{6.1}}& \textcolor{teal}{\textit{6.1}}&\textcolor{teal}{\textit{9.3}} & \textcolor{teal}{\textit{0.4}}& \textcolor{teal}{\textit{2.9}}& \textcolor{teal}{\textit{4.6}}& \textcolor{teal}{\textit{4.1}}& \textcolor{teal}{\textit{12.1}}& \textcolor{teal}{\textit{0.9}}& \textcolor{teal}{\textit{3.9}}& \textcolor{teal}{\textit{3.9}}& \textcolor{teal}{\textit{3.6}} & \textcolor{teal}{\textit{20.7}} & \textcolor{teal}{\textit{0.2}}\\ 
			\midrule[1.2pt]
			GeoTrans~\cite{qin2022geometric} & 71.5 & 78.0 & 83.4 & 26.2 &  2.0 & 48.4 & 65.2 & 74.6 & 51.9 &  5.2 & 62.0 & 72.5 & 75.0 & 97.3 &  0.5  \\
			Generative GeoTrans$^{\rm DINO_{v2}}$ &74.3 & 81.0 & 87.6 & 19.7 &  1.9 & 50.8 & 67.4 & 76.0 & 41.8 &  4.9 & 63.7 & 73.9 & 76.2 & 86.2 &  \textbf{0.4}  \\
			Generative GeoTrans$^{\rm SD}$  & \textbf{77.2} & \textbf{84.0} & \textbf{89.9} & \textbf{16.5} &  \textbf{1.8} & \textbf{51.3} & \textbf{68.7} & \textbf{78.4} & \textbf{35.6} &  \textbf{4.8} & \textbf{65.2} & \textbf{76.1} & \textbf{78.7} & \textbf{71.0} &  \textbf{0.4} \\
			\textit{Improvement} $\uparrow$ & \textcolor{teal}{\textit{5.7}}& \textcolor{teal}{\textit{6.0}}& \textcolor{teal}{\textit{6.5}}& \textcolor{teal}{\textit{9.7}}& \textcolor{teal}{\textit{0.2}}& \textcolor{teal}{\textit{2.9}}& \textcolor{teal}{\textit{3.5}}& \textcolor{teal}{\textit{3.8}}& \textcolor{teal}{\textit{16.3}}& \textcolor{teal}{\textit{0.4}}& \textcolor{teal}{\textit{3.2}}& \textcolor{teal}{\textit{3.6}} & \textcolor{teal}{\textit{3.7}} & \textcolor{teal}{\textit{26.3}} & \textcolor{teal}{\textit{0.1}}\\
			\bottomrule[1.8pt]
	\end{tabular}}
	\label{scannetcompa}
\end{table*}

\begin{table*}[h]
	\centering
	\caption{Comparison of the methods on rotation, translation, and Chamfer distance on \textbf{3DMatch}~\cite{zeng20173dmatch} benchmark dataset.}
	\vspace{-3mm}
	\resizebox{1\textwidth}{!}{
		\begin{tabular}{lccccccccccccccc}
			\toprule[1.8pt]
			&   \multicolumn{5}{c}{\textbf{Rotation} (\textit{deg})} & \multicolumn{5}{c}{\textbf{Translation} (\textit{cm})} & \multicolumn{5}{c}{\textbf{Chamfer} (\textit{mm})} \\
			\cmidrule(lr){2-16} 
			&   \multicolumn{3}{c}{\textbf{Accuracy} $\uparrow$} & \multicolumn{2}{c}{\textbf{Error}$\downarrow$} &  \multicolumn{3}{c}{\textbf{Accuracy} $\uparrow$} & \multicolumn{2}{c}{\textbf{Error}$\downarrow$} & \multicolumn{3}{c}{\textbf{Accuracy} $\uparrow$} & \multicolumn{2}{c}{\textbf{Error}$\downarrow$} \\
			\textbf{Methods}  & 5 & 10 & 45 & Mean & Med. & 5 & 10 & 25 & Mean & Med. & 1 & 5 & 10 & Mean & Med. \\
			\midrule[1.2pt]
			FPFH~\cite{rusu2009fast}  & 69.1 & 82.9 & 91.2 & 15.0 &  3.1 & 25.8 & 53.9 & 75.1 & 37.4 &  9.1 & 52.5 & 74.2 & 79.2 & 57.6 &  0.9  \\
			Lepard~\cite{li2022lepard}  & 84.3 & 91.0 & 94.1 & 11.1 &  2.1 & 43.1 & 75.2 & 88.9 & 21.8 &  5.8 & 72.1 & 88.3 & 90.5 & 45.3 &  0.4  \\
			RegTR~\cite{yew2022regtr}  & 86.2 & 92.1 & 97.2 &  5.7 &  1.6 & 55.0 & 77.6 & 88.9 & 18.8 &  4.6 & 75.4 & 88.2 & 91.3 & 40.0 &  0.3  \\
			RoITr~\cite{yu2023rotation} & 86.3 & 91.1 & 93.8 & 11.1 &  1.6 & 51.2 & 77.4 & 89.1 & 20.5 &  4.9 & 75.2 & 88.5 & 90.6 & 50.1 &  0.4 \\
			\midrule[1.2pt]
			FCGF~\cite{choy2019fully} & 90.4 & 93.7 & 94.8 &  9.4 &  \textbf{1.4} & 53.4 & 79.3 & 91.0 & 19.2 &  4.7 & 76.7 & 90.8 & 92.4 & 40.3 &  \textbf{0.4}  \\
			Generative FCGF$^{\rm DINO_{v2}}$ & 91.5 & 94.3 & 95.3 &  8.5 &  \textbf{1.4} & 53.6 & 79.3 & 91.5 & 18.1 &  \textbf{4.6} & 77.5 & 91.1 & 92.7 & 41.1 &  \textbf{0.4} \\
			Generative FCGF$^{\rm SD}$ & \textbf{94.3} & \textbf{96.7} & \textbf{98.1} &  \textbf{4.5} &  \textbf{1.4} & \textbf{54.3} & \textbf{81.5} & \textbf{93.1} & \textbf{12.5} &  4.7 & \textbf{78.2} & \textbf{92.9} & \textbf{94.6} & \textbf{37.7} &  \textbf{0.4}\\
			\textit{Improvement} $\uparrow$ & \textcolor{teal}{\textit{3.9}}& \textcolor{teal}{\textit{3.0}}& \textcolor{teal}{\textit{3.3}}&\textcolor{teal}{\textit{4.9}} & \textcolor{teal}{\textit{0.0}}& \textcolor{teal}{\textit{0.9}}& \textcolor{teal}{\textit{2.2}}& \textcolor{teal}{\textit{2.1}}& \textcolor{teal}{\textit{6.7}}& \textcolor{teal}{\textit{0.0}}& \textcolor{teal}{\textit{1.5}}& \textcolor{teal}{\textit{2.1}}& \textcolor{teal}{\textit{2.2}} & \textcolor{teal}{\textit{2.6}} & \textcolor{teal}{\textit{0.0}}\\
			\midrule[1.2pt]
			Predator~\cite{huang2021predator} & 85.0 & 91.5 & 94.2 & 10.5 &  2.0 & 42.1 & 72.5 & 87.1 & 22.6 &  5.8 & 71.2 & 85.8 & 88.6 & 45.0 &  0.5  \\
			Generative Predator$^{\rm DINO_{v2}}$ & 88.1 & \textbf{94.8} & 96.9 &  6.2 &  \textbf{1.8} & 44.7 & 73.9 & 88.4 & \textbf{15.5} &  5.6 & 72.4 & 87.7 & 90.8 & \textbf{33.1} &  \textbf{0.4}  \\
			Generative Predator$^{\rm SD}$ & \textbf{88.6} & 94.6 & \textbf{97.0} &  \textbf{5.9} &  1.9 & \textbf{45.7} & \textbf{74.5} & \textbf{89.1} & 15.7 &  \textbf{5.5} & \textbf{73.3} & \textbf{88.3} & \textbf{90.9} & 40.4 &  \textbf{0.4} \\
			\textit{Improvement} $\uparrow$ & \textcolor{teal}{\textit{3.6}}& \textcolor{teal}{\textit{3.1}}& \textcolor{teal}{\textit{2.8}}& \textcolor{teal}{\textit{4.6}}& \textcolor{teal}{\textit{0.2}}& \textcolor{teal}{\textit{3.6}}& \textcolor{teal}{\textit{2.0}}& \textcolor{teal}{\textit{2.0}}& \textcolor{teal}{\textit{7.1}}& \textcolor{teal}{\textit{0.3}}& \textcolor{teal}{\textit{2.1}}& \textcolor{teal}{\textit{2.5}} & \textcolor{teal}{\textit{2.3}} & \textcolor{teal}{\textit{11.9}} & \textcolor{teal}{\textit{0.1}}\\
			\midrule[1.2pt]
			GeoTrans~\cite{qin2022geometric} & 88.9 & 91.8 & 93.3 & 12.0 &  1.4 & 59.8 & 81.0 & 90.1 & 24.6 &  4.0 & 79.2 & 89.0 & 90.6 & 53.3 &  \textbf{0.3}  \\
			Generative GeoTrans$^{\rm DINO_{v2}}$ & 90.2 & 93.2 & 95.2 &  8.9 &  1.4 & 61.0 & \textbf{83.1} & 90.4 & \textbf{16.9} &  \textbf{3.9} & 80.4 & 89.7 & 91.7 & \textbf{36.9} &  \textbf{0.3} \\
			Generative GeoTrans$^{\rm SD}$  & \textbf{91.5} & \textbf{94.3} & \textbf{96.2} &  \textbf{7.6} &  \textbf{1.4} & \textbf{61.3} & 82.9 & \textbf{90.9} & 17.2 &  \textbf{3.9} & \textbf{81.5} & \textbf{90.1} & \textbf{92.3} & 37.3 &  \textbf{0.3} \\
			\textit{Improvement} $\uparrow$ & \textcolor{teal}{\textit{2.6}}& \textcolor{teal}{\textit{2.5}}& \textcolor{teal}{\textit{2.9}}& \textcolor{teal}{\textit{4.4}}& \textcolor{teal}{\textit{0.0}}& \textcolor{teal}{\textit{1.5}}& \textcolor{teal}{\textit{2.1}}& \textcolor{teal}{\textit{0.8}}& \textcolor{teal}{\textit{7.7}}& \textcolor{teal}{\textit{0.1}}& \textcolor{teal}{\textit{2.3}}& \textcolor{teal}{\textit{1.1}} & \textcolor{teal}{\textit{1.7}} & \textcolor{teal}{\textit{16.4}} & \textcolor{teal}{\textit{0.0}}\\
			\bottomrule[1.8pt]
	\end{tabular}}
	\label{3dmatchcomp}
	\vspace{-3mm}
\end{table*}

			\begin{figure*}[t]
	\centering
	\includegraphics[width=\textwidth]{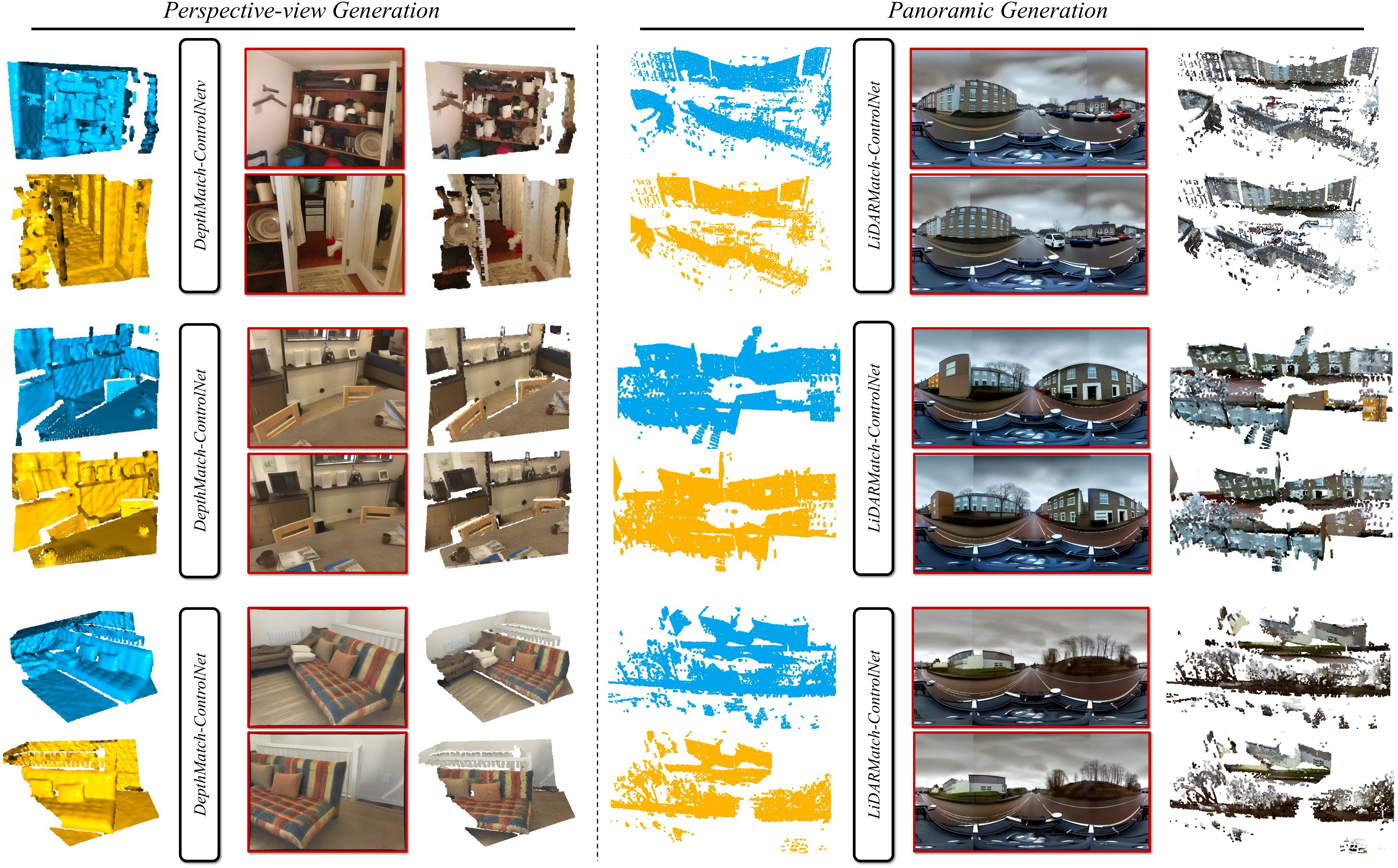}
	\caption{The visualization of the generated perspective-view/panoramic RGB image pairs and the formed color source and target point clouds using the \textit{DepthMatch-ControlNet} (\textit{left}) and \textit{LiDARMatch-ControlNet} (\textit{right}). 
	}
	\label{visxyz}
\end{figure*}

	\subsection{Geometric-Color Fused Point Descriptor}\label{fuse}
	In this section, we focus on how to enhance the geometric representations of point clouds with the free-lunch color information from generated perspective/panoramic source and target images, denoted as  $\tilde{\mathbf{I}}_{\mathcal{P}}$ and $\tilde{\mathbf{I}}_{\mathcal{Q}}$. We  provide two geometric-color fusion schemes:
	
	\noindent \textbf{Zero-Shot Geometric-Color Feature Fusion.} Inspired by the powerful RGB representations of large vision models, we utilize them to directly extract zero-shot semantic features from the generated images. Specifically, we employ two widely-used vision foundation models: DINOv2~\cite{oquab2023dinov2} and Stable Diffusion~\cite{rombach2022high} for image encoding, achieving corresponding feature maps. These feature maps are then projected into the point cloud space using the camera intrinsic matrix, yielding pointwise color descriptors: $\{\mathbf{f}^{rgb}_{\mathbf{p}_i}\}$ and $\{\mathbf{f}^{rgb}_{\mathbf{q}_i}\}$ for both source and target point clouds.
	Finally, we apply a simple weighted concatenation to combine the RGB descriptors with the geometric descriptors, resulting in fused descriptors:
	\begin{equation}\label{fuseform}
	\tilde{\mathbf{f}}_{\mathbf{p}_i} = [\omega \cdot \mathbf{f}^{geo}_{\mathbf{p}_i}; (1 - \omega) \cdot \operatorname{PCA}(\mathbf{f}^{rgb}_{\mathbf{p}_i})]\in\mathbb{R}^{d_{geo}+d_{rgb}},
	\end{equation}
	where $[\cdot; \cdot]$ denotes the feature concatenation operator; $\omega\in[0, 1]$ is the fusion weight and $\mathbf{f}_{\mathbf{p}_i}^{geo}$ represents the geometric descriptors; $\operatorname{PCA}(\cdot)$ denotes the principal component analysis function to compress the feature dimension of the color descriptor to fit that of the geometric descriptor. The same fusion scheme is also applied to the target point clouds. Notably, this zero-shot geometric-color fusion approach is general and can be applied to a variety of geometric descriptors, whether traditional or deep descriptors.
	
	\noindent \textbf{XYZ-RGB Fusion.} This fusion scheme directly projects the generated source and target RGB images into the point cloud space. The resulting RGB values are then concatenated with the point coordinates of the point clouds, forming 6D color source and target point clouds, as shown in Fig.~\ref{visxyz}. These color point clouds are subsequently used as inputs to the color point cloud registration method, like ColorPCR~\cite{mu2024colorpcr}, for 3D registration.

		\begin{figure}[t]
	\centering
	\includegraphics[width=\columnwidth]{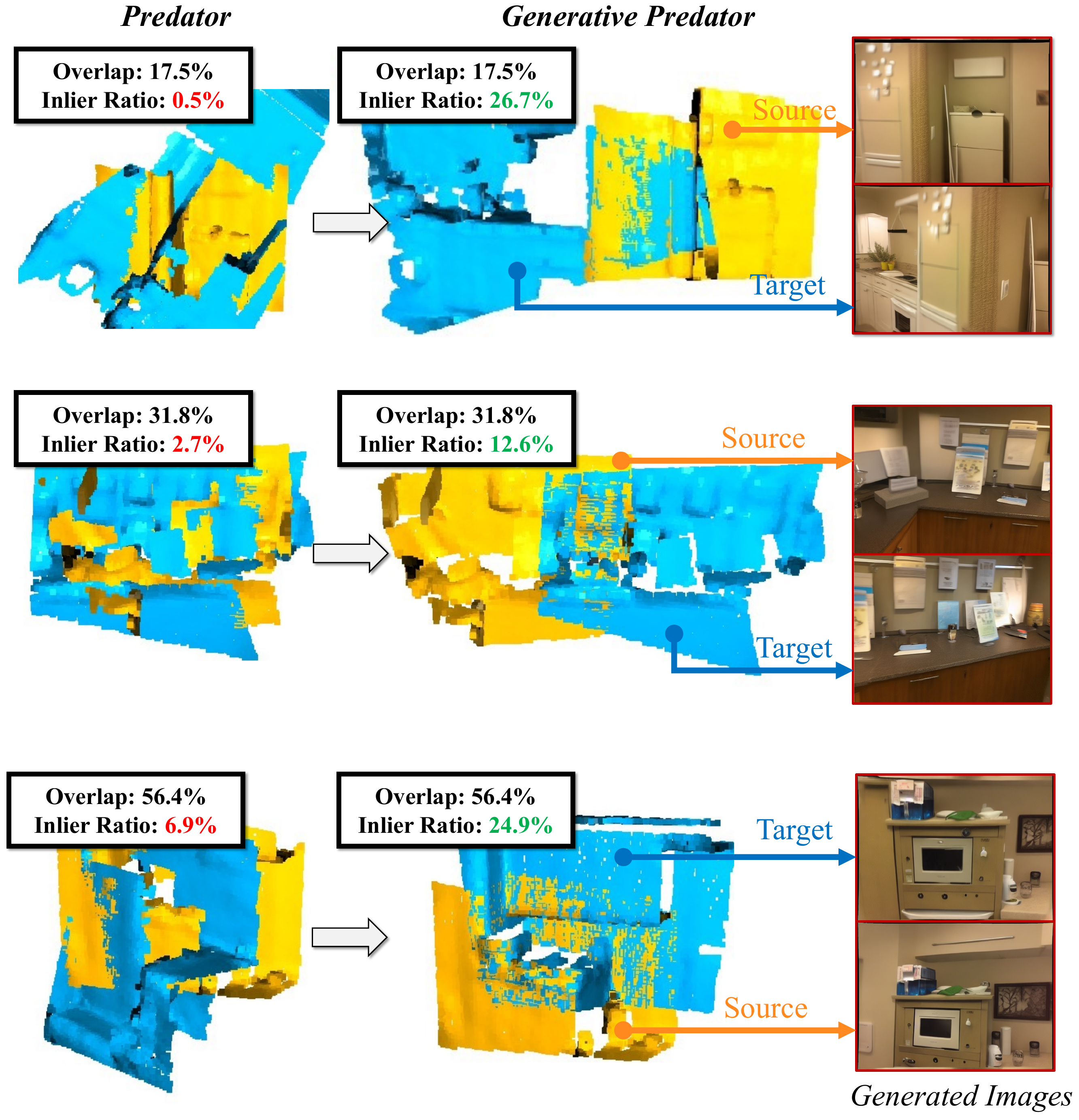}
	\vspace{-7mm}
	\caption{Qualitative comparions between Predator and Generative Predator on \textbf{3DMatch}~\cite{zeng20173dmatch}. The original Predator struggles with accurate registration. By contrast, the  {Generative Predator}, enhanced with our generated color information, achieves significantly better alignment.
	}
	\label{3dimg2}
	\vspace{-3mm}
\end{figure}

	    	\begin{table*}[t]
		\centering
		\caption{Comparison of the methods on FMR, IR, and RR on \textbf{Dur360BEV}~\cite{yuan2025dur360bev} dataset with point cloud pairs separated by at least 5 or 10 meters.}
		\vspace{-3mm}
		\resizebox{0.85\textwidth}{!}{
			\begin{tabular}{lccc|ccc}
				\toprule[1.8pt]
                 & \multicolumn{3}{c}{\textbf{$\geq$ 5m Apart}} & \multicolumn{3}{c}{\textbf{$\geq$ 10m Apart}}\\
                 \cmidrule(lr){2-7} 
				\textbf{Methods}  & FMR (\%) $\uparrow$ & IR (\%) $\uparrow$ & RR (\%) $\uparrow$ & FMR (\%) $\uparrow$ & IR (\%) $\uparrow$ & RR (\%) $\uparrow$ \\
				\midrule[1.2pt]
				PARE-Net~\cite{yao2024pare} & 93.3 & 24.3 & 69.6 & 76.5 & 13.5 & 56.7 \\
                CoFiNet~\cite{yu2021cofinet} & 100.0 & 34.2 & 99.4 & 95.6 & 20.5 & 90.5 \\
				\midrule[1.2pt]
				FPFH~\cite{rusu2009fast}  & 75.1 & 6.4 & 11.9  & 14.6 & 3.8 & 0.7  \\
				Generative FPFH$^{\rm DINOv2}$ & 100.0 & 18.8 & 86.7  & 84.9 & 8.5 & 30.6 \\
				Generative FPFH$^{\rm SD}$ & 100.0 & 21.8 & 88.6 & 86.7 & 9.0 & 31.0 \\
				\textit{Improvement} $\uparrow$ & \textcolor{teal}{\textit{24.9}} & \textcolor{teal}{\textit{15.4}} & \textcolor{teal}{\textit{76.7}}  & \textcolor{teal}{\textit{72.1}} & \textcolor{teal}{\textit{5.2}} & \textcolor{teal}{\textit{30.3}} \\
				\midrule[1.2pt]
				FCGF~\cite{choy2019fully}  & 100.0 & 41.3 & 96.3 & 100.0 & 22.8 & 85.3 \\
				 Generative FCGF$^{\rm DINO_{v2}}$ & 100.0 & 43.6 & 97.6 & 99.9 & 24.0 & 89.3  \\
				Generative FCGF$^{\rm SD}$ & 100.0 & 44.2 & 98.7 & 99.9 & 23.0 & 91.2\\
				\textit{Improvement} $\uparrow$ & \textcolor{teal}{\textit{0.0}} & \textcolor{teal}{\textit{2.9}} & \textcolor{teal}{\textit{2.4}} & \textcolor{bittersweet}{\textit{0.1}} & \textcolor{teal}{\textit{1.2}} & \textcolor{teal}{\textit{5.9}} \\
				\midrule[1.2pt]
				Predator~\cite{huang2021predator} & 99.9 & 23.0 & 95.8 & 97.3 & 12.6 & 57.9 \\
				 Generative Predator$^{\rm DINO_{v2}}$ & 100.0 & 31.2 & 98.1 & 99.0 & 16.2 & 79.0 \\
				Generative Predator$^{\rm SD}$ & 100.0 & 33.3 & 98.4 & 98.7 & 16.1 & 77.7 \\
				\textit{Improvement} $\uparrow$ & \textcolor{teal}{\textit{0.1}} & \textcolor{teal}{\textit{10.3}} & \textcolor{teal}{\textit{2.6}} & \textcolor{teal}{\textit{1.4}} & \textcolor{teal}{\textit{3.5}} & \textcolor{teal}{\textit{19.8}} \\
                \midrule[1.2pt]
                GeoTrans~\cite{qin2022geometric}  & 100.0 & 62.9 & 99.8 &  98.3 & 40.1 & 94.9 \\
				 Generative GeoTrans$^{\rm DINO_{v2}}$ & 100.0 & 63.4 & 100.0 & 99.9 & 41.1 & 95.8  \\
				Generative GeoTrans$^{\rm SD}$ & 100.0 & 63.8 & 99.9 & 99.8 & 41.1 & 95.5\\
				\textit{Improvement} $\uparrow$ & \textcolor{teal}{\textit{0.0}} & \textcolor{teal}{\textit{0.7}} & \textcolor{teal}{\textit{0.2}} & \textcolor{teal}{\textit{1.6}} & \textcolor{teal}{\textit{1.0}} & \textcolor{teal}{\textit{0.9}} \\
				\bottomrule[1.8pt]
		\end{tabular}}
		\label{outdoorcompare}
		\vspace{-3mm}
	\end{table*}

	\section{Experiments}
	\subsection{Experimental Setting}
	\noindent\textbf{Implementation Details.} 
During the few-shot fine-tuning stage, we randomly select 3,000 sample pairs from the ScanNet training set~\cite{dai2017scannet} for model fine-tuning. Following the default fine-tuning configuration of ControlNet~\cite{Zhang_2023_ICCV}, we adopt the AdamW optimizer~\cite{loshchilov2017decoupled} with a learning rate of 1e-5 and set the training epoch to 10. 
The code for this project is implemented in PyTorch, and all experiments are conducted on a server equipped with an Intel i5 2.2 GHz CPU and a TITAN RTX GPU.

		\begin{figure}[t]
	\centering
	\includegraphics[width=\columnwidth]{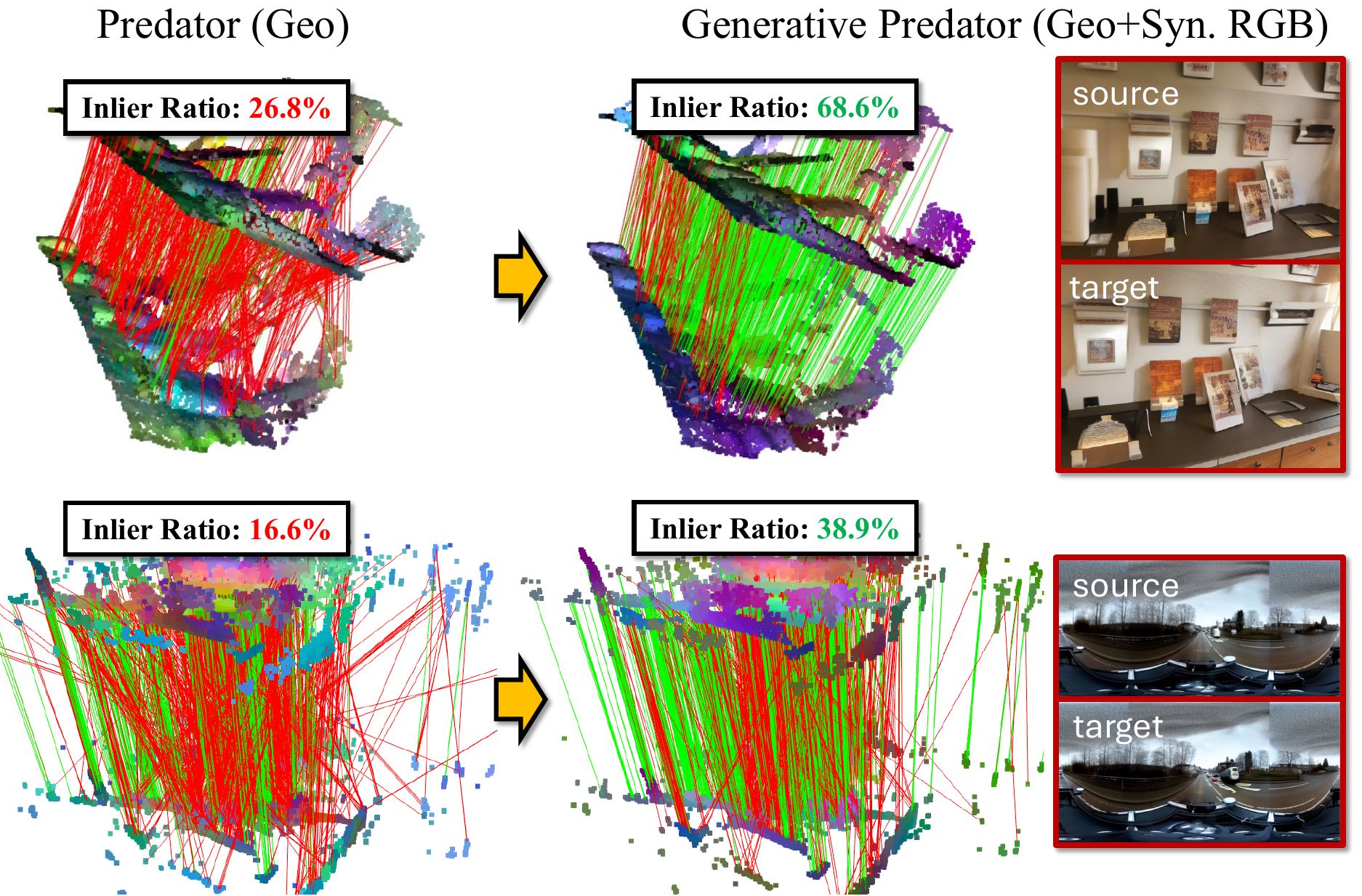}
	\caption{Correspondence visualization of Predator and Generative Predator on \textbf{3DMatch} dataset~\cite{zeng20173dmatch} (\textit{top}) and \textbf{Dur360DEV} dataset~\cite{yuan2025dur360bev} (\textit{bottom}). Benefitting from the rich RGB information from generated perspective-view/panoramic images, our Generative-Predator descriptors presents higher inlier ratios than the geometry-only Predator descriptors.
	}
	\label{vis3dreg}
\end{figure}

	\noindent\textbf{Evaluation Metric.} Following~\cite{el2021unsupervisedr,yuan2023pointmbf}, we use rotation error, translation error, and Chamfer error, including the accuracy across varying thresholds and mean/median errors, for performance evaluation on depth camera-based 3D registration. Instead, for LiDAR-based 3D registration, we follow the evaluation protocol proposed in~\cite{huang2021predator,qin2022geometric}, adopting three metrics: Inlier Ratio (\textit{IR}), Feature Matching Recall (\textit{FMR}), and Registration Recall (\textit{RR}) for evaluation. 
	
	\subsection{Comparison with Existing Methods} \label{comparequa}
		\noindent\textbf{Evaluation on ScanNet.} We first test our \textit{generative point cloud registration} paradigm on the depth camera-based 3D registration. We perform model evaluation on a widely-used, large-scale indoor benchmark dataset, ScanNet~\cite{dai2017scannet}, where the point clouds are derived from depth maps captured by depth sensors. We follow the official data split to divide this dataset into the training, validation, and testing subsets, and construct view pairs by sampling image pairs that are 50 frames apart. 
		Compared to the 20-frame separation used in~\cite{el2021unsupervisedr,yuan2023pointmbf}, our approach with a 50-frame separation further reduces the overlap ratio (i.e., lower overlap), thereby increasing the registration difficulty.

		We compare our method against with one representative traditional descriptor: FPFH \cite{rusu2009fast}, one scene-level end-to-end registration network: RegTR~\cite{yew2022regtr}, and five deep descriptors: FCGF~\cite{choy2019fully}, Predator~\cite{huang2021predator}, GeoTrans~\cite{qin2022geometric}, Lepard~\cite{li2022lepard}, and RoITr~\cite{yu2023rotation}. 
		Here, we integrate FCGF, Predator, and GeoTrans into our  \textit{generative point cloud registration} framework, resulting in corresponding color-enhanced variants: {Generative FCGF}, {Generative Predator}, and {Generative GeoTrans} for method evaluation. 
		We adopt RANSAC-50k as the pose estimator for FPFH, Lepard and RoITr, and select SC2PCR~\cite{chen2022sc2}, RANSAC-50k and LGR for (Generative) FCGF, (Generative) Predator, and (Generative) GeoTrans~\cite{qin2022geometric}, respectively, to validate the robustness of our generative 3D registration paradigm across different pose estimators.  Table~\ref{scannetcompa} demonstrates that enhanced by the free-lunch color information generated by our \textit{DepthMatch-ControlNet}, all generative versions of FCGF, Predator, and GeoTrans achieve significant performance improvements, such as $6.9\%\uparrow$ of Generative FCGF on 45$^\circ$@Rotation metric. These confirm the generality and effectiveness of our framework. Additionally, we find that compared to the DINOv2 image encoding, Stable-Diffusion (\textit{SD}) encoding can capture more discriminative representations and achieve higher precisions. 
		
	\noindent\textbf{Evaluation on 3DMatch.} We next evaluate our method on 3DMatch~\cite{zeng20173dmatch}, another widely-used benchmark dataset for single-view 3D registration. We follow~\cite{el2021unsupervisedr,yuan2023pointmbf} as in ScanNet to produce the pairwise samples. 
	Also, we increase the view separation from 20 to 40, resulting in point cloud pairs with lower overlap to increase the registration challenge.  
Table~\ref{3dmatchcomp} shows that integrating FCGF, Predator, and GeoTrans into our \textit{generative point cloud registration} framework consistently improves their performance, further validating the effectiveness of our proposed paradigm. Additional qualitative comparisons and correspondence visualizations are presented in Fig.~\ref{3dimg2} and Fig.~\ref{vis3dreg}, respectively. 

				\begin{table*}[ht]
		\centering
		\caption{Ablation studies on \textbf{3DMatch}~\cite{zeng20173dmatch} dataset. (*) denotes the default configuration.}
		\vspace{-3mm}
		\resizebox{1\textwidth}{!}{
			\begin{tabular}{lccccccccccccccc}
				\toprule[1.8pt]
				&   \multicolumn{5}{c}{\textbf{Rotation} (\textit{deg})} & \multicolumn{5}{c}{\textbf{Translation} (\textit{cm})} & \multicolumn{5}{c}{\textbf{Chamfer} (\textit{mm})} \\
				\cmidrule(lr){2-16} 
				&   \multicolumn{3}{c}{\textbf{Accuracy} $\uparrow$} & \multicolumn{2}{c}{\textbf{Error}$\downarrow$} &  \multicolumn{3}{c}{\textbf{Accuracy} $\uparrow$} & \multicolumn{2}{c}{\textbf{Error}$\downarrow$} & \multicolumn{3}{c}{\textbf{Accuracy} $\uparrow$} & \multicolumn{2}{c}{\textbf{Error}$\downarrow$} \\
				\textbf{Methods}  & 5 & 10 & 45 & Mean & Med. & 5 & 10 & 25 & Mean & Med. & 1 & 5 & 10 & Mean & Med. \\
				\midrule[1.2pt] 
				FCGF & 90.4 & 93.7 & 94.8 &  9.4 &  \textbf{1.4} & 53.4 & 79.3 & 91.0 & 19.2 &  \textbf{4.7} & 76.7 & 90.8 & 92.4 & 40.3 &  \textbf{0.4}  \\
				Generative FCGF$^{\rm SD}$ (geo) & {92.4} & {96.1} & {97.8} &  {5.2} &  {1.5} & 53.3 & 79.2 & 91.7 & 13.5 &  4.8 & 75.6 & 91.1 & 93.1 & \textbf{35.8} &  \textbf{0.4} \\
				Generative FCGF$^{\rm SD}$ (geo + tex)  &  \textbf{94.3} & \textbf{96.7} & \textbf{98.1} &  \textbf{4.5} &  \textbf{1.4} & \textbf{54.3} & \textbf{81.5} & \textbf{93.1} & \textbf{12.5} &  \textbf{4.7} & \textbf{78.2} & \textbf{92.9} & \textbf{94.6} & 37.7 &  \textbf{0.4}\\
				\midrule[1.2pt]
				Generative FCGF$^{\rm SD}$ (zero-shot) & {92.4} & {96.1} & 97.3 &  5.4 &  {1.5} & \textbf{54.3} & {80.3} & {92.4} & {13.0} &  \textbf{4.6} & \underline{77.3} & {92.1} & {94.0} & \textbf{33.7} &  \textbf{0.4}\\
				Generative FCGF$^{\rm SD}$ (finetuning) & \textbf{94.3} & \textbf{96.7} & \textbf{98.1} &  \textbf{4.5} &  \textbf{1.4} & \textbf{54.3} & \textbf{81.5} & \textbf{93.1} & \textbf{12.5} &  4.7 & \textbf{78.2} & \textbf{92.9} & \textbf{94.6} & 37.7 &  \textbf{0.4} \\
				\cdashline{1-16}[3pt/3pt]
				Finetune (\#samples=1000) & 93.5 & 96.8 &  \textbf{98.1} &  4.6 &   \textbf{1.4} & 54.2 & 80.5 & 92.5 & 12.4 &  4.7 &  \textbf{78.2} & 92.5 &  \textbf{94.7} & 32.8 &   \textbf{0.4} \\
				Finetune (\#samples=3000)* & \textbf{94.3} & 96.7 &  \textbf{98.1} &  4.5 &   \textbf{1.4} &  \textbf{54.3} &  \textbf{81.5} &  \textbf{93.1} & 12.5 &  4.7 &  \textbf{78.2} &  \textbf{92.9} & 94.6 & 37.7 &   \textbf{0.4}  \\
				Finetune (\#samples=5000) & 93.6 &  \textbf{97.2} & 98.0 &   \textbf{4.4} &  1.5 & 54.0 & 80.8 & 92.7 &  \textbf{11.9} &   \textbf{4.5} & 77.7 &  \textbf{92.9} & 94.2 &  \textbf{32.3} &   \textbf{0.4} \\
				\midrule[1.2pt]
				ColorPCR~\cite{mu2024colorpcr} & 79.9 & 84.6 & 88.9 & 16.5 &   \textbf{1.8} &  \textbf{48.3} & 69.6 & 82.2 & 41.8 &   \textbf{5.2} & 66.6 & 80.6 & 83.3 & 81.1 &  0.5  \\
				Generative ColorPCR &  \textbf{83.6} &  \textbf{89.8} &  \textbf{93.2} &  \textbf{12.0} &  1.9 & 47.3 &  \textbf{73.3} &  \textbf{86.7} &  \textbf{28.3} &  5.3 &  \textbf{70.3} &  \textbf{85.7} &  \textbf{88.2} &  \textbf{59.1} &   \textbf{0.4}\\
				\textit{Improvement} $\uparrow$ &  \textcolor{teal}{\textit{3.7}}& \textcolor{teal}{\textit{5.2}}& \textcolor{teal}{\textit{4.3}}& \textcolor{teal}{\textit{4.5}}& \textcolor{bittersweet}{\textit{0.1}}& \textcolor{bittersweet}{\textit{1.0}}& \textcolor{teal}{\textit{3.7}}& \textcolor{teal}{\textit{4.5}}& \textcolor{teal}{\textit{13.5}}& \textcolor{bittersweet}{\textit{0.1}}& \textcolor{teal}{\textit{3.7}}& \textcolor{teal}{\textit{5.1}} & \textcolor{teal}{\textit{4.9}} & \textcolor{teal}{\textit{22.0}} & \textcolor{teal}{\textit{0.1}}\\
				\midrule[1.2pt]
				Color feat. dim. $d_{rgb}=16$ &90.8 & {93.9} & \textbf{96.3} &  \textbf{7.6} &  \textbf{1.4} & \textbf{62.2} & \textbf{83.4} & 90.7 & 18.1 &  3.9 & \textbf{81.5} & 89.7 & 91.7 & 41.8 &  \textbf{0.3} \\
				Color feat. dim. $d_{rgb}=32$& \textbf{91.5} & \textbf{94.3} & {96.2} &  \textbf{7.6} &  \textbf{1.4} & 61.3 & 82.9 & 90.9 & \textbf{17.2} &  3.9 & \textbf{81.5} & 90.1 & 92.3 & \textbf{37.3} &  \textbf{0.3}  \\
				Color feat. dim. $d_{rgb}=64$*& \textbf{91.5} & \textbf{94.3} & {96.2} &  \textbf{7.6} &  \textbf{1.4} & 61.3 & 82.9 & 90.9 & \textbf{17.2} &  3.9 & \textbf{81.5} & 90.1 & 92.3 & \textbf{37.3} &  \textbf{0.3}  \\
				Color feat. dim. $d_{rgb}=128$&  {91.0} & \textbf{94.3} & 96.0 &  8.1 &  \textbf{1.4} & 62.0 & \textbf{83.4} & \textbf{91.6} & 18.1 &  \textbf{3.8} & \textbf{81.5} & \textbf{90.4} & \textbf{92.4} & 41.2 &  \textbf{0.3}  \\
				\midrule[1.2pt] 
				Fusion weight $\omega=0.0$& 88.3 & 93.3 & 96.6 &  6.9 &  1.6 & 53.7 & 77.0 & 86.5 & 20.8 &  4.7 & 74.6 & 85.9 & 88.5 & 50.5 &  0.4 \\
				Fusion weight $\omega=0.25$& 90.7 & \textbf{95.5} & \textbf{97.2} &  \textbf{6.1} &  1.5 & 57.9 & 81.5 & 89.8 & \textbf{16.5} &  4.2 & 78.6 & 89.2 & 92.0 & 40.4 &  \textbf{0.3}  \\
				Fusion weight $\omega=0.50$*& \textbf{91.5} & 94.3 & 96.2 &  7.6 &  \textbf{1.4} & \textbf{61.3} & \textbf{82.9} & \textbf{90.9} & 17.2 &  \textbf{3.9} & \textbf{81.5} & \textbf{90.1} & \textbf{92.3} & \textbf{37.3} &  \textbf{0.3}  \\
				Fusion weight $\omega=0.75$&  89.2 & 92.7 & 93.9 & 10.9 &  \textbf{1.4} & 60.5 & 81.7 & 90.6 & 22.6 &  4.0 & 79.5 & 89.3 & 91.4 & 49.5 &  \textbf{0.3}  \\
				Fusion weight $\omega=1.0$& 89.0 & 91.8 & 93.3 & 12.0 &  \textbf{1.4} & 59.9 & 81.1 & 90.2 & 24.6 &  4.0 & 79.3 & 89.1 & 90.8 & 53.3 &  \textbf{0.3} \\
				\bottomrule[1.8pt]
		\end{tabular}}
		\label{ablationstudy}
\vspace{-3mm}
	\end{table*}

\begin{figure}[t]
	\centering
	\includegraphics[width=\columnwidth]{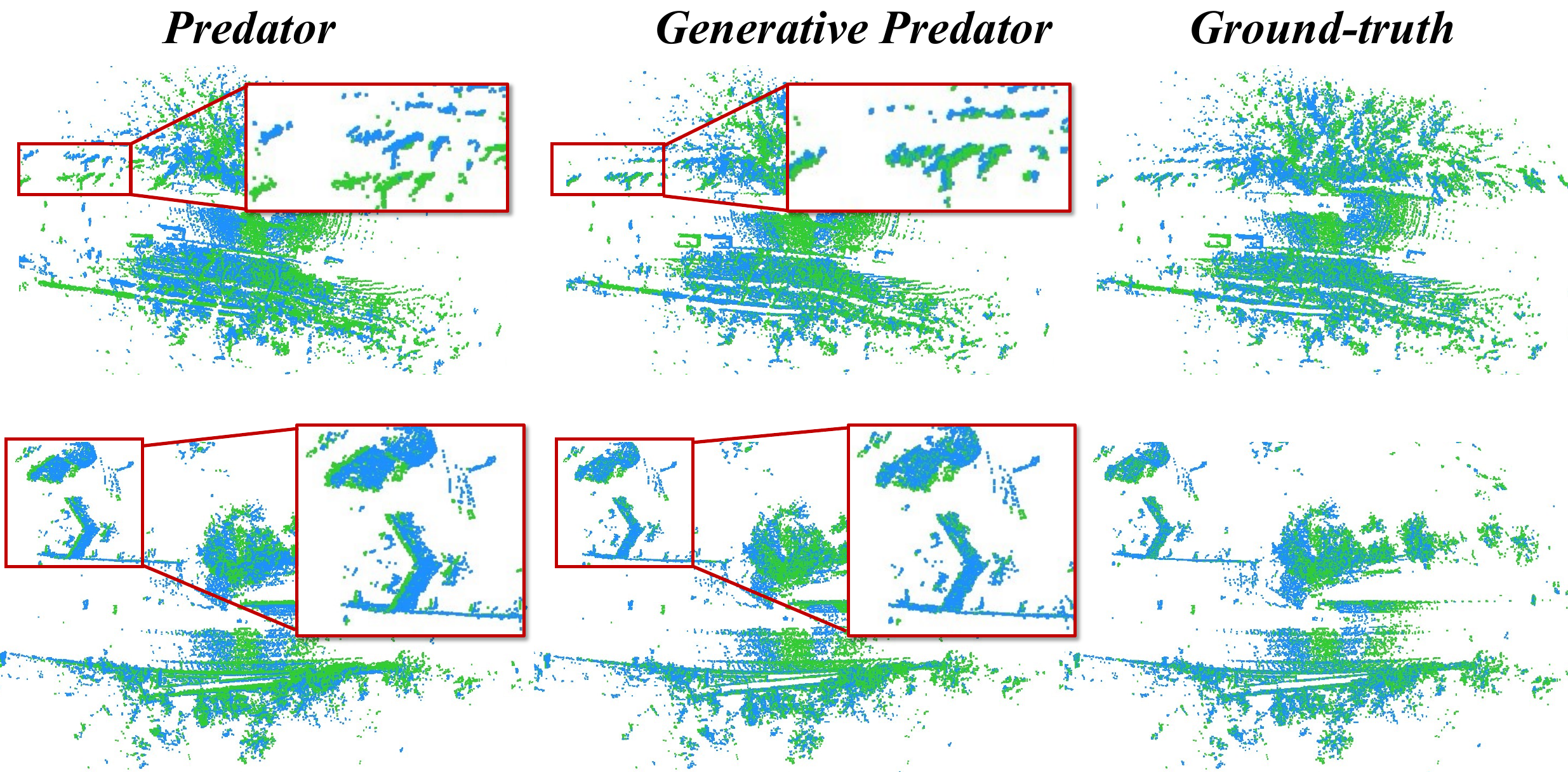}
			\vspace{-7mm}
	\caption{Qualitative comparions between Predator and Generative Predator on \textbf{Dur360BEV}~\cite{yuan2025dur360bev} benchmark dataset. 
	}
	\label{vis3dreg22}
	\vspace{-2mm}
\end{figure}

\noindent\textbf{Evaluation on Dur360BEV.}  
We evaluate the effectiveness of our \textit{generative point cloud registration} approach on LiDAR-based 3D registration using outdoor LiDAR-scanned driving scenarios from the Dur360BEV dataset~\cite{yuan2025dur360bev}. Following the official data protocol, we use 14,767 frames with timestamps 0--11,509 and 13,150--16,406 as the training set, and the remaining 1,640 frames with timestamps 11,510--13,149 as the testing set. 
Consistent with~\cite{choy2019fully}, we select only point cloud pairs that are at least 5 or 10 meters apart in spatial distance for training and testing, and apply ICP to refine the ground-truth rigid transformations. 
We compare our method with a traditional descriptor, FPFH~\cite{rusu2009fast}, and five deep learning-based descriptors: PARE-Net~\cite{yao2024pare}, CoFiNet~\cite{yu2021cofinet}, FCGF~\cite{choy2019fully}, Predator~\cite{huang2021predator}, and GeoTrans~\cite{qin2022geometric}. Note that methods such as Lepard, RegTR, and RoITr are not included in the comparison, as their pretrained models are not available for LiDAR settings. 
We integrate FPFH, FCGF, Predator, and GeoTrans into our  \textit{generative point cloud registration} framework, resulting in corresponding color-enhanced variants: {Generative FPFH}, {Generative FCGF}, {Generative Predator}, and Generative GeoTrans for method evaluation. 
As shown in Table~\ref{outdoorcompare}, the proposed generative variants tend to consistently achieve performance gains over their respective baselines, benefitting from the high-quality panoramic image generation provided by our \textit{LiDARMatch-ControlNet}. Notably, Generative FPFH achieves a Registration Recall (\textit{RR}) gain of 76.7\%, while Generative Predator reaches an Inlier Ratio (\textit{IR}) gain of 10.3\%. These impressive accuracy gains highlight the strong effectiveness of our generative paradigm. 
More qualitative comparisons are also provided in Fig.~\ref{vis3dreg} and Fig.~\ref{vis3dreg22}.

	\subsection{Ablation Studies and Analysis}	
	\noindent\textbf{Effectiveness of Match-ControlNet.} 
	We first evaluate the performance contribution of our developed \textit{Match-ControlNet}: 
	\textbf{(i)} The top block of Table~\ref{ablationstudy} demonstrates that, compared to using generated image pairs with only 2D-3D geometric consistency (\textit{geo}), incorporating both 2D-3D geometric consistency and cross-view texture consistency (\textit{geo+tex}) through our \textit{Match-ControlNet} results in higher registration accuracy. This improvement is due to the additional benefit of consistent textures and colors, which further facilitate accurate correspondence identification. Additionally, we observe that the generated images with only 2D-3D geometric consistency can also bring performance gain in some criteria. We attribute it to that DINOv2 and Stable Diffusion can extract powerful semantic representations, mitigating the feature inconsistency of correspondences caused by the texture difference and thereby aiding correspondence identification. Furthermore, we visualize the generated image pair for given source and target point clouds in Fig.~\ref{visxyz}. It shows that our \textit{Match-ControlNet} is capable of producing high-quality image pairs with consistent 2D-3D geometry and cross-view texture.

	\noindent\textbf{Zero-Shot vs Finetuning.} We further investigate the performance of \textit{DepthMatch-ControlNet} in both zero-shot and finetuned settings. As shown in the second block of Table~\ref{ablationstudy}, both approaches yield substantial improvements over FCGF. Moreover, because the finetuned \textit{DepthMatch-ControlNet} benefits from task-specific training, it consistently achieves higher registration accuracy than the zero-shot version. Notably, even few-shot finetuning with as few as 1K samples yields clear performance gains. Increasing the number of finetuning samples (e.g., to 3K or 5K) provides additional improvements; however, models trained on 3K or 5K samples show comparable registration accuracy in practice. Hence, we adopt 3K samples as our default finetuning configuration. 
	
	\noindent\textbf{Zero-Shot Geometric-Color Feature Fusion.} We next conduct ablation studies on the zero-shot geometric-color feature fusion described in Eq.~\ref{fuseform}. As shown in the fourth block of Table~\ref{ablationstudy}, Generative GeoTrans exhibits varying registration performance under different color feature dimensions, \(d_{rgb} \in \{16, 32, 64, 128\}\). We observe that a very small color feature dimension (e.g., \(d_{rgb} = 16\)) degrades performance due to limited semantic representational capacity, while excessively large dimensions do not yield significant performance gains. Therefore, to balance inference efficiency with registration precision, we set \(d_{rgb} = 64\) as our default setting. 
	Additionally, in the fifth block of Table~\ref{ablationstudy}, we investigate performance variations with different fusion weights \(\omega \in \{0.0, 0.25, 0.50, 0.75, 1.0\}\), where a larger \(\omega\) places more emphasis on the geometric descriptors (see Eq.~\ref{fuseform}). Our results indicate that both overly high \(\omega\) (which overemphasizes geometry) and overly low \(\omega\) (which overemphasizes color) lead to degraded registration accuracy. By contrast, a balanced weight (e.g., \(\omega = 0.50\)) achieves higher performance. As a result, we adopt \(\omega = 0.50\) as our default hyperparameter configuration.

\begin{figure}[t]
		\centering
		\includegraphics[width=\columnwidth]{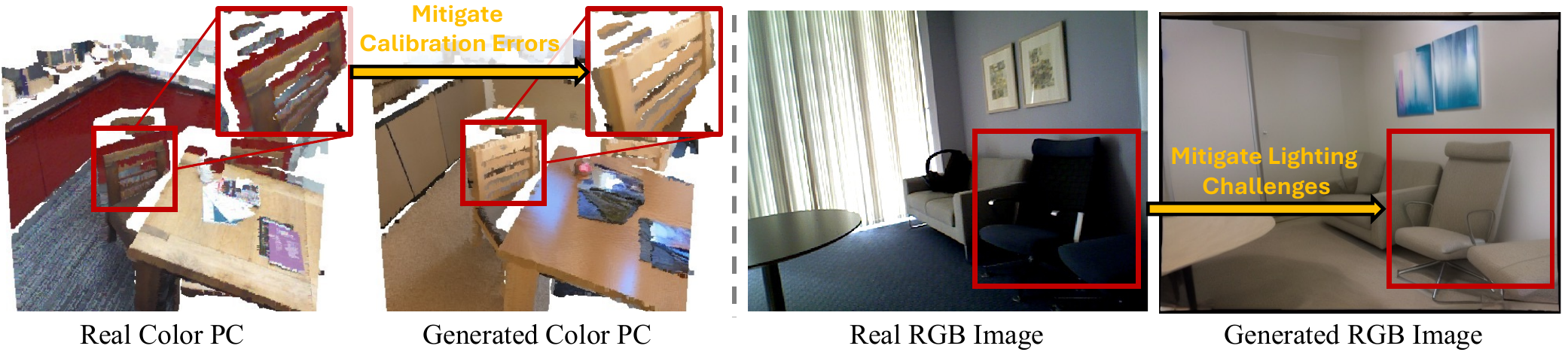}
		\vspace{-7mm}
		\caption{Our \textit{Match-ControlNet} effectively mitigates calibration errors and lighting challenges commonly encountered in real-world RGB-D data, thereby improving the matching precision of color point cloud registration methods. For example, Generative ColorPCR using \textit{synthetic RGB data} even outperforms the original ColorPCR trained with \textit{real RGB data}, as reported in the third block of Table~\ref{ablationstudy}.
		}
		\label{error}
		\vspace{-5mm}
	\end{figure}

\noindent\textbf{XYZ-RGB Fusion.} We finally evaluate the effectiveness of the XYZ-RGB fusion (see Sec.~\ref{fuse}). We replace the real color point clouds (with actual RGB values) used by ColorPCR~\cite{mu2024colorpcr} with our generated color point clouds (with synthesized RGB values), forming \textbf{Generative ColorPCR} for 3D matching. The third block in Table~\ref{ablationstudy} demonstrates that, on the 3DMatch dataset, Generative ColorPCR with the synthesized color even outperforms the original ColorPCR with the real color. This advantage is attributed not only to the high-quality pairwise image generation provided by our \textit{Match-ControlNet}, but also to several key benefits of our generated XYZ-RGB data over real XYZ-RGB data: 
\textbf{(i)} Mitigating calibration errors:
As shown in Fig.~\ref{error} (\textit{left}), some real RGB-D data would suffer from calibration errors, which may lead to misalignment in the colored point clouds. By contrast, our framework, benefiting from the powerful 2D-3D consistency generation ability, effectively reduces such calibration errors, producing higher-quality colored point clouds and enabling more accurate matching. 
\textbf{(ii)} Mitigating lighting challenges:
Fig.~\ref{error} (\textit{right}) shows that RGB images from real-world conditions can degrade under poor lighting, negatively impacting RGB-D point cloud matching performance. By contrast, our \textit{generative point cloud registration} framework potentially produces images with consistent lighting conditions, independent of real-world illumination variations, thereby enhancing the overall robustness of our method to lighting challenges.

	\section{Conclusion}
	We have introduced a novel 3D registration paradigm, \emph{generative point cloud registration}, which effectively leverages advanced 2D generative models to augment geometry-only 3D registration.
	To this end, we developed \emph{DepthMatch-ControlNet} and \textit{LiDARMatch-ControlNet}, two matching-specific variants of ControlNet designed to synthesize paired perspective-view/panoramic RGB images for both source and target point clouds.
	By integrating depth/range map-conditioned generation from ControlNet, coupled conditional denoising, and coupled prompt guidance, these generated RGB image pairs preserve both 2D-3D geometric consistency and cross-view texture consistency, thereby facilitating high-quality 3D matching.
	Notably, our generative framework is general and can be incorporated into a variety of registration methods to improve their registration performance.
	Extensive experiments on both depth camera-based/LiDAR-based 3D registration tasks demonstrate the effectiveness of the proposed framework.

	\section*{ACKNOWLEDGMENTS}    This research is supported by MOE AcRF Tier 1 Grant of Singapore (RG12/22) and  by the RIE2025 Industry Alignment Fund – Industry Collaboration Projects (IAF-ICP) (Award I2301E0026), administered by A*STAR, as well as supported by Alibaba Group and NTU Singapore through Alibaba-NTU Global e-Sustainability CorpLab (ANGEL).

\ifCLASSOPTIONcaptionsoff
  \newpage
\fi

\bibliographystyle{IEEEtran}
\bibliography{main}

\begin{IEEEbiography}[{\includegraphics[width=11in,height=1.2525in,clip,keepaspectratio]{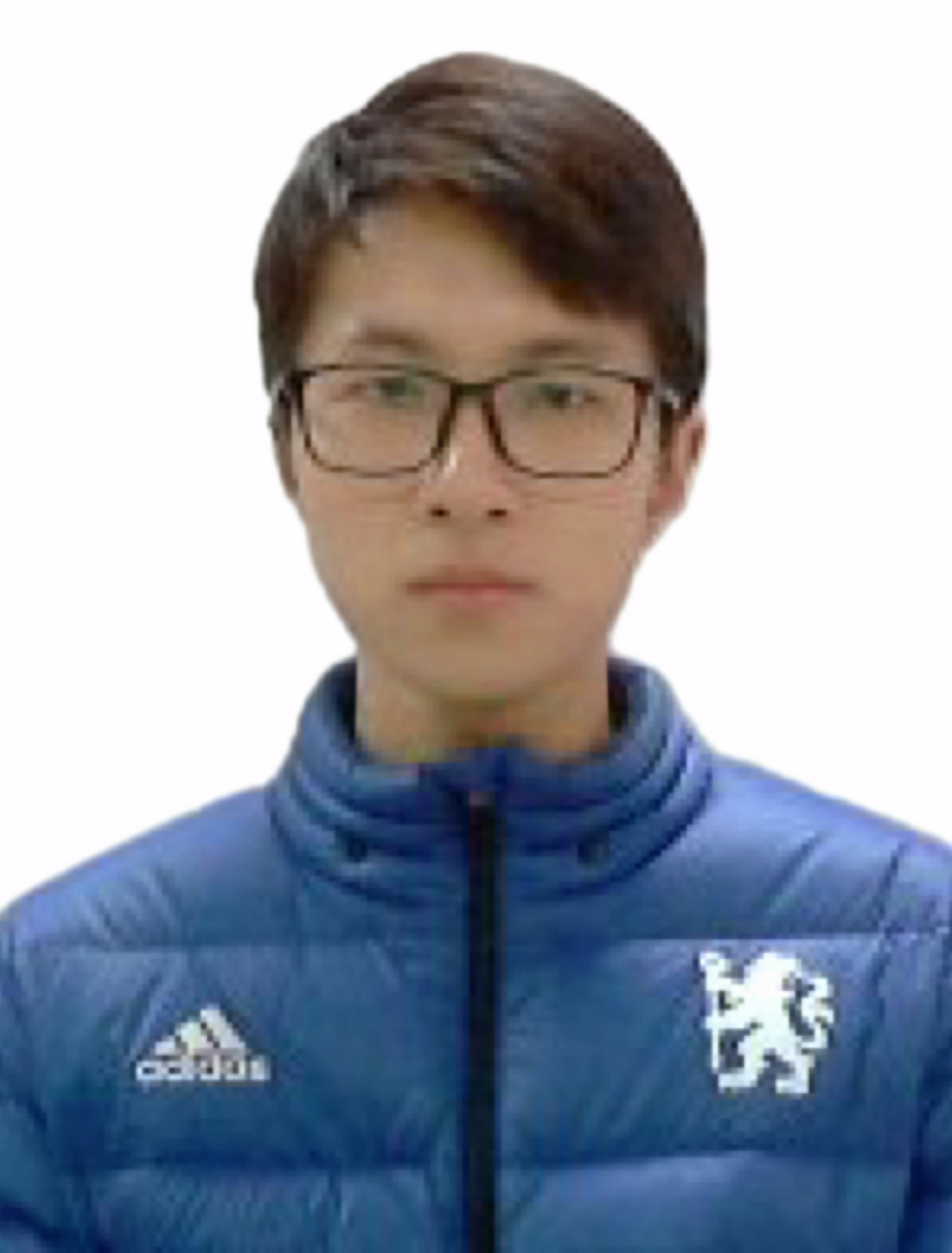}}]{Haobo Jiang} is currently a Postdoctoral Researcher at School of Computer Science and Engineering, Nanyang Technological University, Singapore. He received the Ph.D. degree in control science and engineering in the School of Computer Science and Engineering, Nanjing University of Science and Technology (NUST), Nanjing, China. His current research interests focus on reinforcement learning and 3D computer vision, including robotic control and planning, 3D point cloud registration, 3D object tracking and pose estimation, and 3D scene reconstruction.  
\end{IEEEbiography}

\begin{IEEEbiography}[{\includegraphics[width=11in,height=1.2525in,clip,keepaspectratio]{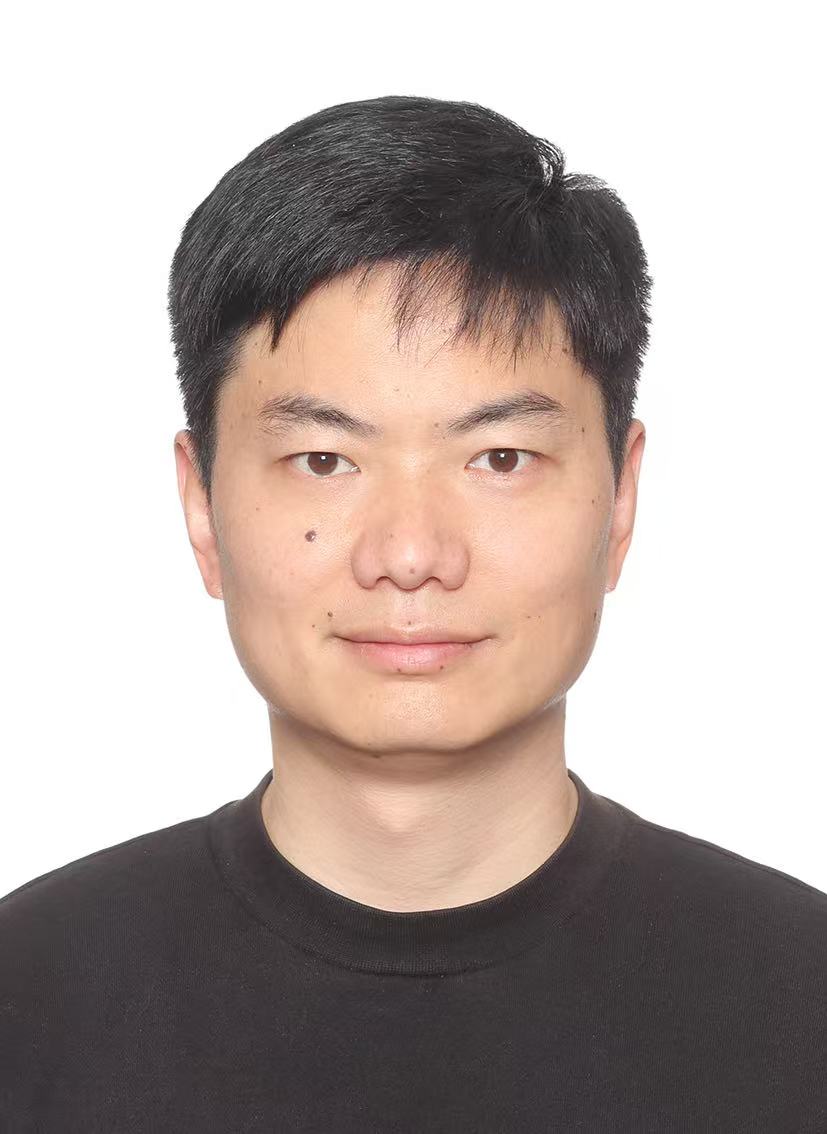}}]{Liang Yu} received the Ph.D. degree in photogrammetry and remote sensing from Wuhan University in 2008. He is currently a senior staff engineer with Alibaba Cloud, and leading the delivery of AI solutions to enterprises. Prior to Alibaba, he did postdoctoral research with the National University of Singapore, the National Center for Supercomputing Applications of the University of Illinois at Urbana-Champaign, and Singapore-MIT Alliance for Research and Technology. His research interest has covered a wide range of smart city topics such as geospatial data analysis, multimodal model, LLM agent, etc. 
\end{IEEEbiography}

\begin{IEEEbiography}[{\includegraphics[width=11in,height=1.2525in,clip,keepaspectratio]{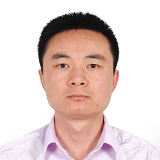}}]{Jin Xie} is currently a professor in the School of Intelligence Science and Technology, Nanjing University, China. He received his Ph.D degree from the Department of Computing, The Hong Kong Polytechnic University, in 2012. He was a research scientist at New York University Abu Dhabi from 2013 to 2017. Prior to joining Nanjing University in 2023, he was a professor in the Department of Computer Science and Engineering at Nanjing University of Science and Technology, China.  His research interests include machine learning, computer vision, computer graphics and robotics. His current research focus is on 3D computer vision and its applications on autonomous driving and robotic manipulation. He has authored/co-authored over 50 papers in well-known journals/conferences such as IEEE TPAMI, IJCV, CVPR, ICCV, ECCV and NeurIPS. He has served as a reviewer for IEEE TPAMI, TIP, TNNLS, TMM, CVPR, ICCV and ECCV. He was a special issue chair for Asian Conference on Pattern Recognition 2017 and a guest editor for Pattern Recognition. He received the best paper award for Asian Conference on Pattern Recognition 2021.
\end{IEEEbiography}

\begin{IEEEbiography}
	[{\includegraphics[width=11in,height=1.2525in,clip,keepaspectratio]{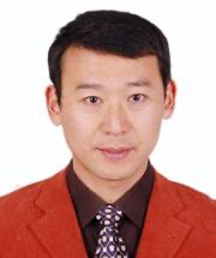}}]{Jian Yang}
	received the PhD degree from Nanjing University of Science and Technology (NUST) in 2002, majoring in pattern recognition and intelligence systems. From 2003 to 2007, he was a Postdoctoral Fellow at the University of Zaragoza, Hong Kong Polytechnic University and New Jersey Institute of Technology, respectively. From 2007 to present, he is a professor in the School of Computer Science and Technology of NUST. Currently, he is also a visiting distinguished professor in the College of Computer Science of Nankai University. He is the author of more than 300 scientific papers in pattern recognition and computer vision. His papers have been cited over 40000 times in the Scholar Google. His research interests include pattern recognition and computer vision. Currently, he is/was an associate editor of Pattern Recognition, Pattern Recognition Letters, IEEE Trans. Neural Networks and Learning Systems, and Neurocomputing. He is a Fellow of IAPR. 
\end{IEEEbiography}

\begin{IEEEbiography}[{\includegraphics[width=11in,height=1.2525in,clip,keepaspectratio]{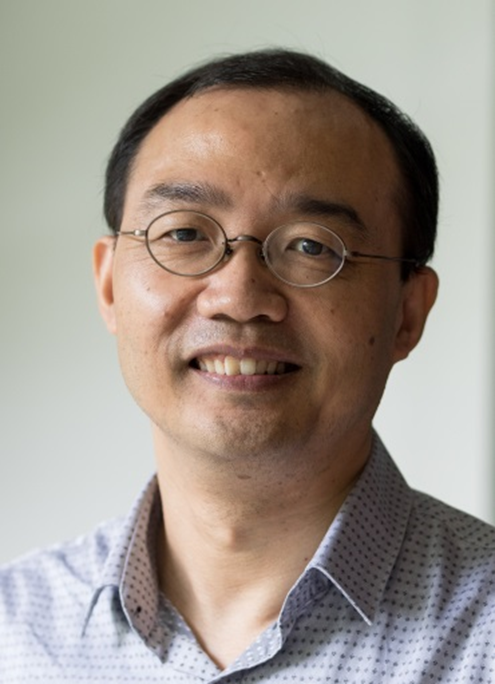}}]{Jianmin Zheng} is a professor
in the College of Computing and
Data Science at Nanyang Technological
University, Singapore. He received his B.S. and
Ph.D. degrees from Zhejiang University,
China. His recent research
focuses on T-spline technologies,
intelligent geometric processing,
AI for part design, 3D printing,
3D graphics, visualization, social
robots and virtual humans. He is
on the editorial board of several
journals including Computer-Aided
Design, IEEE Computer Graphics and Applications, The Visual Computer, and
Computers and Graphics.
He is an SMA fellow.
\end{IEEEbiography}

\end{document}